\documentclass[acmsmall]{acmart}
\usepackage{hyperref}
\usepackage{graphicx}
\usepackage{multirow}
\usepackage{amsmath}
\usepackage{xcolor}
\usepackage{booktabs}
\usepackage{caption}
\usepackage{subcaption}
\AtBeginDocument{%
  \providecommand\BibTeX{{%
    \normalfont B\kern-0.5em{\scshape i\kern-0.25em b}\kern-0.8em\TeX}}}

\setcopyright{acmcopyright}
\copyrightyear{2022}
\acmYear{2022}
\acmDOI{XXXXXXX.XXXXXXX}

\acmJournal{TIST}
\acmVolume{0}
\acmNumber{0}
\acmArticle{0}
\acmMonth{0}

\begin{document}

\title{Multitask Balanced and Recalibrated Network for Medical Code Prediction}

\author{Wei~Sun}
\affiliation{%
  \institution{Aalto University}
  \city{Espoo}
  \country{Finland}
}
\email{wei.sun@aalto.fi}

\author{Shaoxiong~Ji}
\affiliation{%
  \institution{Aalto University}
  \city{Espoo}
  \country{Finland}
}
\email{shaoxiong.ji@aalto.fi}

\author{Erik~Cambria}
\affiliation{%
  \institution{Nanyang Technological University}
  \country{Singapore}
}
\email{cambria@ntu.edu.sg}

\author{Pekka~Marttinen}
\affiliation{%
  \institution{Aalto University}
  \city{Espoo}
  \country{Finland}
}
\email{pekka.marttinen@aalto.fi}

\renewcommand{\shortauthors}{Sun, et al.}

\begin{abstract}
Human coders assign standardized medical codes to clinical documents generated during patients' hospitalization, which is error-prone and labor-intensive. 
Automated medical coding approaches have been developed using machine learning methods such as deep neural networks. 
Nevertheless, automated medical coding is still challenging because of complex code association, noise in lengthy documents, and the imbalanced class problem.
We propose a novel neural network called Multitask Balanced and Recalibrated Neural Network to solve these issues.
Significantly, the multitask learning scheme shares the relationship knowledge between different coding branches to capture the code association.
A recalibrated aggregation module is developed by cascading convolutional blocks to extract high-level semantic features that mitigate the impact of noise in documents.
Also, the cascaded structure of the recalibrated module can benefit the learning from lengthy notes.
To solve the imbalanced class problem, we deploy the focal loss to redistribute the attention on low and high-frequency medical codes.
Experimental results show that our proposed model outperforms competitive baselines on a real-world clinical dataset MIMIC-III.
\end{abstract}

\ccsdesc[500]{Applied computing~Health care information systems}
\ccsdesc[300]{Computing methodologies~Natural language processing}
\ccsdesc[100]{Applied computing~Document management and text processing}

\keywords{Medical Code Prediction, Multitask Learning, Imbalanced Class Problem, Balanced and Recalibrated Network}

\maketitle
\section{Introduction}

Professional doctors write discharge summaries based on different kinds of clinical notes such as diagnosis reports, prescriptions, and treatment procedure documents. 
Health institutes annotate these notes with standardized medical codes to facilitate information acquisition and management. 
International Classification of Disease~(ICD)\footnote{\url{https://www.who.int/standards/classifications/classification-of-diseases}}, one of the most widely used medical coding systems, is maintained by the World Health Organization~(WHO).
ICD coding system converts disease, pathology reason, symptom, and signs into standard ICD codes, which is helpful in various medical-related services, including insurance reimbursement~\cite{park2000accuracy}, statistical data analysis, and clinical decision support~\cite{horng2017creating}.
Since medical codes annotation by human is error-prone~\cite{o2005measuring} and labor intensive~\cite{sun2021multitask}, a surge of feature engineering-based machine learning~\cite{perotte2014diagnosis, koopman2015automatic} and deep learning~\cite{vu2020label,sun2021multitask, ji2021medical, li2020icd} approaches have been proposed for automating the medical coding task. 

However, the automatic medical coding task is still challenging, mainly due to the following three aspects.

\begin{figure}[htbp]
\centering
\begin{subfigure}[]{0.4\textwidth}
    \includegraphics[width=\textwidth]{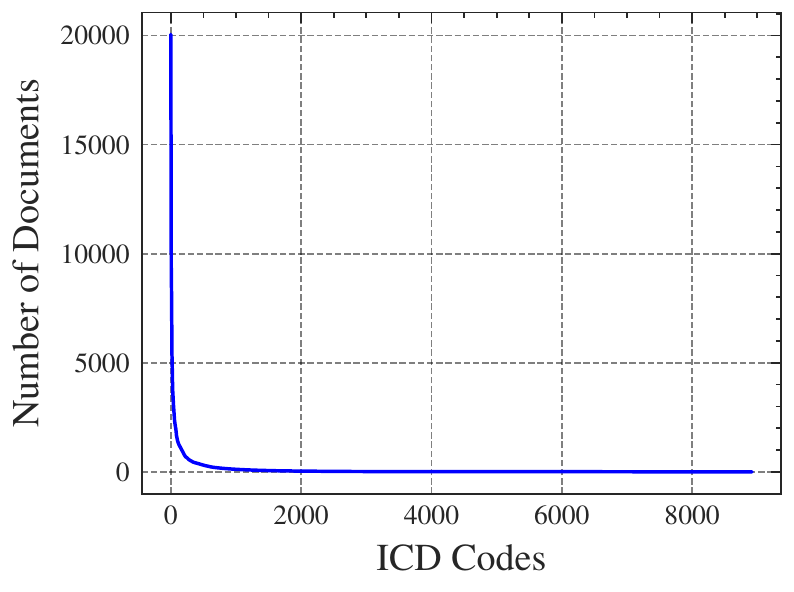}
    \caption{ICD}
    \label{fig:dist_ICD}
\end{subfigure}
\quad
\begin{subfigure}[]{0.4\textwidth}
    \includegraphics[width=\textwidth]{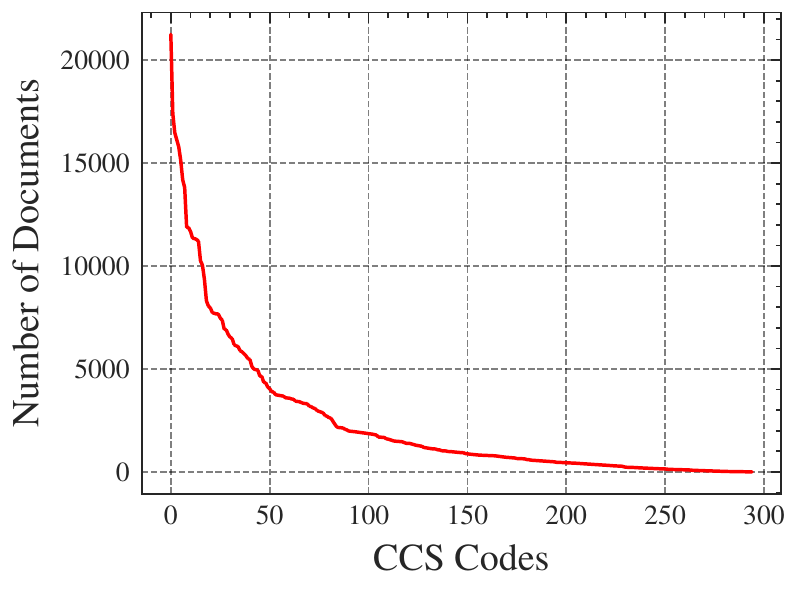}
    \caption{CCS}
    \label{fig:dist_CCS}
\end{subfigure}
\caption{Distributions of ICD and CCS codes in the MIMIC-III dataset. We omit a multitude of x-axis tick labels for better demonstration. The ICD codes are sorted by code frequency, which decreases from left to right.}
\label{fig:Distribution}
\end{figure}

\textbf{Imbalanced Class Problem:}
Many medical coding datasets, such as the third version of the Medical Information Mart for Intensive Care (MIMIC-III), suffer from a severe imbalanced class problem. 
We take the MIMIC-III dataset as an example, the frequency distribution of ICD codes and CCS codes demonstrated in Figure~\ref{fig:dist_ICD} and Figure~\ref{fig:dist_CCS}, respectively.
The main reason is that people encounter some diseases such as ``Hyperlipidemia'' and ``Type II diabetes'' that are more frequent than other low-frequency diseases, such as ``Angioneurotic edema'' and ``Quadriplegia''.
Model learning will be biased toward frequent labels if we train models on an imbalanced dataset without any re-balance strategies.
Therefore, re-balancing the learning of low- and high-frequency code can improve the performance of medical coding approaches.

Some advanced ICD coding approaches mitigate imbalanced class problem based on the nature of ICD codes.
For example, the \textbf{La}bel \textbf{at}tention model~(LAAT)~\cite{vu2020label} proposed the hierarchical joint learning mechanism to capture the structure of ICD codes to help predict the low-frequency ICD codes.

Inspired by the imbalanced problem between easy and hard samples in the object detection task, we regard low and high-frequency medical codes as hard-classified and easy-classified samples.
Our framework leverages the focal loss~\cite{lin2017focal} to re-balance the attention on low- and high-frequency code by adjusting the loss weights dynamically.

\textbf{Code Association:}
There exist some connections between medical codes.
For example, in the ICD taxonomy system, ``427.31'' and ``427.89'' represent ``Atrial fibrillation'' and ``Other specified cardiac dysrhythmias'', respectively, which can be classified into ``Dysrhythmia''.
It is challenging to capture associations between medical codes to facilitate the automatic medical coding task.
Existing ICD coding models such as \textbf{Multi}-Filter \textbf{Res}idual \textbf{C}onvolutional \textbf{N}eural \textbf{N}etwork~(MultiResCNN)~\cite{li2020icd} and \textbf{C}onvolutional \textbf{A}ttention for \textbf{M}ulti-\textbf{L}abel classification~(CAML)~\cite{mullenbach2018explainable} did not take medical codes association into consideration. 

\begin{figure}[htbp]
\centering
\includegraphics[width=0.6\linewidth]{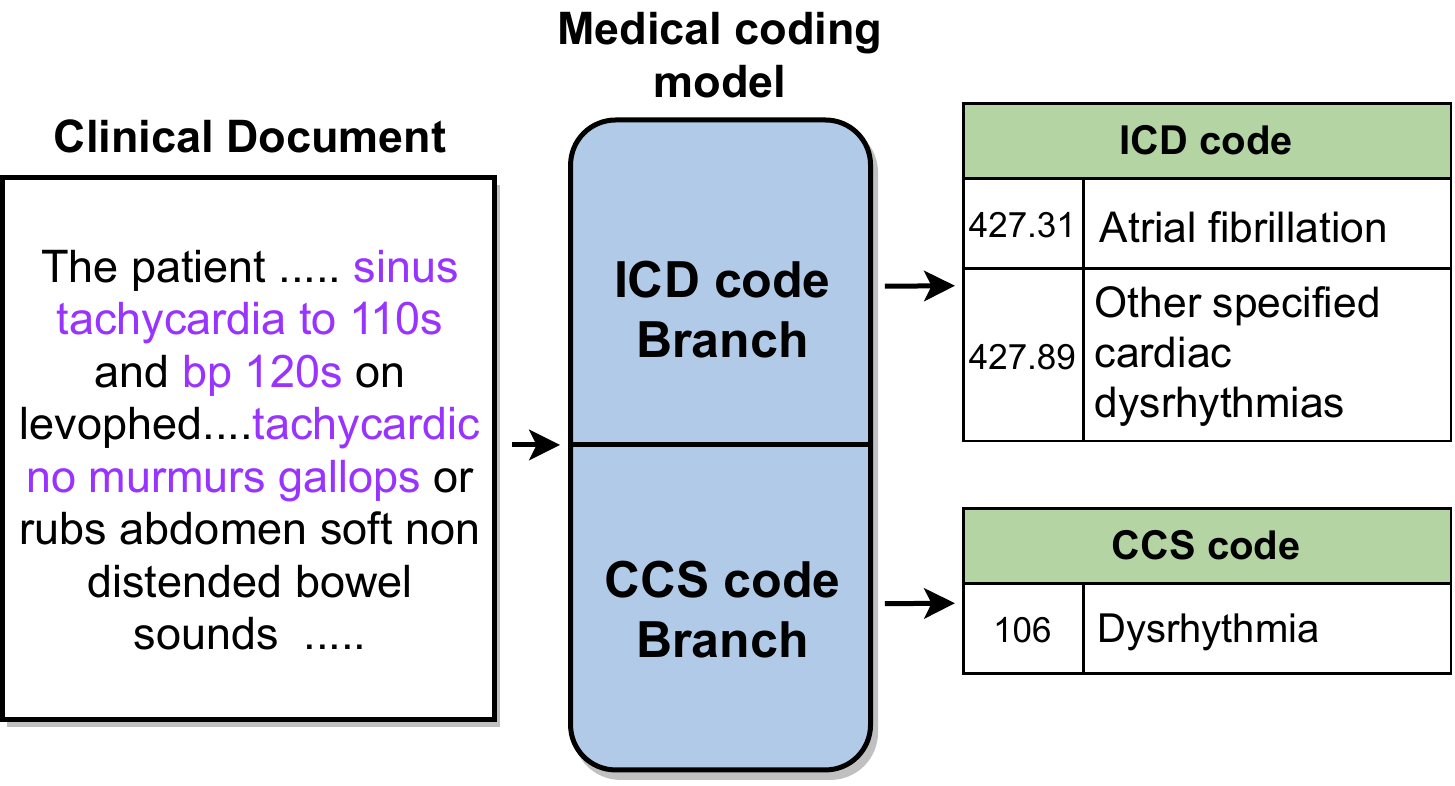}
\caption{An example of multitask coding with ICD and CCS codes}
\label{fig:example}
\end{figure}

We design a multitask learning scheme to effectively transfer ICD code association knowledge from the auxiliary task branch.
The Clinical Classifications Software (CCS) system maps high-dimensional ICD codes into low-dimensional CCS codes. 
The code projection system is based on the pre-defined medical knowledge of code association information provided by the Healthcare Cost and Utilization Project~(HCUP).
The object is to jointly train two medical coding branches, i.e., ICD and CCS coding branches.
The CCS coding branch is an auxiliary task to transfer ICD code association information.
Figure~\ref{fig:example} shows an example of two-branch multitask medical coding, where two ICD codes of ``427.31''~(Atrial fibrillation) and ``427.89''~(Other specified cardiac dysrhythmias) map to the same CCS codes ``106''~(Dysrhythmia).

\begin{figure}[htbp]
\centering
\includegraphics[width=0.6\linewidth]{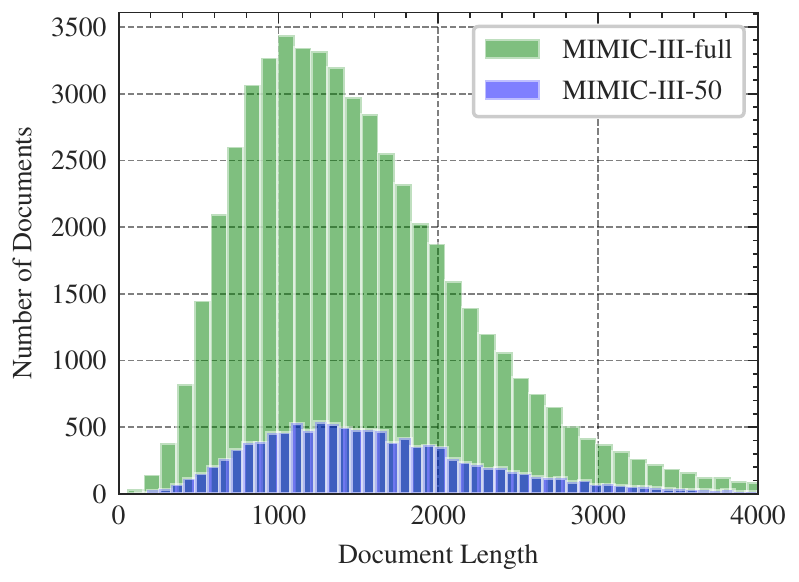}
\caption{\textcolor{black}{Distributions of clinical document lengths in the MIMIC-III datasets.}}
\label{fig:doc_len}
\end{figure}

\textbf{Noisy and Lengthy Document:}
Clinical documents contain noisy information, including error spelling and incoherent information, affecting the representation learning from text.
Learning rich and robust document features is required to provide reliable medical coding results. 
\textcolor{black}{Moreover, discharge summaries in the MIMIC-III dataset are extremely long, $96.3$\% and $97.2$\% documents' lengths exceed $512$ tokens in the MIMIC-III-full and MIMIC-III-50 datasets.
Figure~\ref{fig:doc_len} shows that most documents' lengths are on the interval of $[1000, 2000]$, which are quite long compared with texts in other domains such as the IMDB movie review dataset~\cite{maas2011learning}.}
To improve feature learning from clinical documents, we design a module called \textbf{R}ecalibrated \textbf{A}ggregation \textbf{M}odule~(RAM).
RAM suppresses the noise in the clinical notes by injecting contextually enhanced document features. 
The cascaded convolution structure of RAM provides the model with the capability to better deal with lengthy clinical documents.

To address the three problems mentioned above, we propose a novel framework called \textbf{M}ultitask b\textbf{A}lanced and \textbf{R}ecalibrated \textbf{N}etwork~(MARN) for medical codes prediction.
This paper is an extension of our previous work that proposed a model called MT-RAM~\cite{sun2021multitask} to jointly train two different medical coding branches and achieve competitive improvement on overall evaluation metrics.
Our additional work includes the efforts to balance the learning from frequent and infrequent codes, the extension of experiments on a full-code dataset with improved performance, and further analysis on the problem of imbalanced class and code association.
Our proposed MARN model combines multitask learning, bidirectional gated recurrent unit~(BiGRU), RAM, label-aware attention mechanism, and focal loss.
We summarize our main contributions as follows:
\begin{itemize}
    \item This paper deals with the imbalanced class problem and leverages focal loss to dynamically redistribute the weight between low- and high-frequency codes.  
    \item We utilize a multitask learning scheme to jointly train two medical coding systems with different granularities for capturing code associations.
    \item \textbf{R}ecalibrated \textbf{A}ggregation \textbf{M}odule~(RAM) is designed to refine textual features extracted from lengthy and noisy clinical notes.
    \item Experimental results show strong performance of our model across different evaluation metrics on a widely used dataset MIMIC-III in a comparison against several strong baseline models.
\end{itemize}

Our paper is organized as follows: Section~\ref{sec:related} introduces related work; 
Section~\ref{sec:method} describes our proposed MARN model; 
Section~\ref{sec:exp} conducts a series of experiments and explores components of the MARN;
Section~\ref{sec:future} discusses the future direction;
Section~\ref{sec:conclusion} concludes the paper.

\section{Related Work}
\label{sec:related}

\textbf{Automatic Medical Coding}
Automatic medical coding is a challenging but essential task in medical text mining~\cite{perotte2014diagnosis}.
Early automatic medical coding models mainly depend on complicated hand-craft document features.
Larkey and Croft~\cite{larkey1996combining} designed an ICD code classifier by conflating a potpourri of machine learning components, including K-nearest neighbor, relevance feedback, and Bayesian independence classifiers.
Perotte et al.~\cite{perotte2014diagnosis} proposed two ICD coding models, i.e., a flat and a hierarchy-based SVM classifier.
Comparison experiments demonstrated the superiority of the hierarchical-based model because it can capture the hierarchical structure of ICD codes, which can benefit the prediction of ICD codes.

Recent years have witnessed the advances in deep learning approaches.
Mullenbach et al.~\cite{mullenbach2018explainable} proposed \textbf{C}onvolutional \textbf{A}ttention network for \textbf{M}ulti-\textbf{L}abel classification~(CAML) for automatic ICD coding.
Li and Yu~\cite{li2020icd} designed a \textbf{Mul}ti-Filter \textbf{Res}idual \textbf{C}onvolutional \textbf{N}eural \textbf{N}etwork (MultiResCNN).
Ji et al.~\cite{ji2020dilated} developed a dilated convolutional network.
Dong et al.~\cite{dong2021explainable} devised a hierarchical label-wise attention network.
Thanh et al.~\cite{vu2020label} developed a label attention model (LAAT) to predict ICD codes.
Xie et al.~\cite{xie2019ehr} designed a multi-scale feature attention and structured knowledge graph propagation~(MSATT-KG), which is the combination of a densely connected convolutional neural network~(CNN), multi-scale feature attention and graph convolutional neural networks.
The densely connected CNN generates the n-gram features, and the multi-scale feature attention captures the most informative n-gram document features. 
Also, the MSAAT-KG uses a graph convolutional neural network to obtain the hierarchical structure of ICD codes and the semantics of each ICD code.

\begin{figure}[htbp]
\centering
\includegraphics[width=0.6\linewidth]{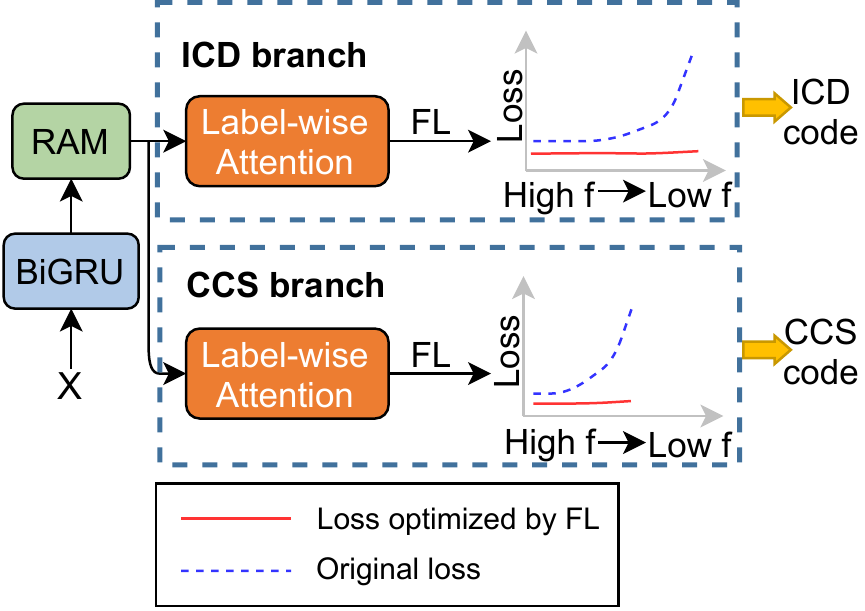}
\caption{Overall architecture of MARN. $f$ represents the code frequency. After utilizing focal loss, the losses of low- and high-frequency medical codes can be balanced. }
\label{fig:MARN}
\end{figure}

\textbf{Imbalanced Class Problem}
The imbalanced class problem is founded on the different distributions of class labels in the data set~\cite{chawla2004special}.
The distribution of the label space demonstrates a head-tail shape, where head and tail labels represent high and low-frequency labels, respectively. 
Most multi-label data sets suffer from serious imbalanced class problem~\cite{tahir2012multilabel}.
The conventional way to alleviate the imbalanced class problem divides instances into majority~(classes with notable instances) and minority groups~(classes with few instances)~\cite{charte2015addressing}.
Then, sampling algorithms are applied to reduce the number of majority samples or increase the number of minority samples.
In past decades, many re-sampling approaches have been proposed, such as Random Over-Sampling~(ROS), SMOTE~\cite{chawla2002smote} and Random Under-Sampling(RUS)~\cite{kotsiantis2003mixture}.
However, reconstructing the data from a medical data set is expensive and inefficient. 
Therefore, most automatic coding works tend to deal with the imbalanced class problem by leveraging the hierarchical structure of ICD codes. 
For example, LAAT~\cite{vu2020label} proposed the joint hierarchical mechanism to deal with the imbalanced class problem.
On the other hand, we determine to handle the imbalanced class problem by dynamically redistributing the loss weights between low- and high-frequency medical codes.

\textbf{Multitask Learning}
Multitask learning~(MTL) is inspired by human activities, where people can utilize experience from other tasks to prompt the learning process of the new task~\cite{zhang2017survey}.
MTL contributes to the information communication between related tasks by sharing parameters and increases the training efficiency~\cite{chandra2016evolutionary, yosinski2014transferable}.
Additionally, MTL can alleviate the over-fitting problem by regularizing the learned model parameters to improve the model's generalization ability for each task branch~\cite{liu2019multi}.
Many MTL-based approaches have been proposed to verify the feasibility and effectiveness of applying the MTL scheme to medical natural language processing (NLP) tasks such as medical named entity recognition~\cite{zhao2019neural, chowdhury2018multitask}, clinical information extraction~\cite{suk2016deep, bi2008improved}  and morality prediction~\cite{si2019deep}.
However, only a few narratives deploy the multitask learning scheme on the automatic medical coding task.
Interian et al.~\cite{interian2020multitask} studied two different healthcare tasks, i.e., medical code prediction and morality prediction, to perform multitask training.

\section{Method}
\label{sec:method}
This section introduces our proposed model, \textbf{M}ultitask b\textbf{A}alanced and \textbf{R}ecalibrated \textbf{N}etwork~(MARN),  which consists of a multitask learning scheme, \textbf{R}ecalibrated \textbf{A}ggregation \textbf{M}odule (RAM) and optimization with the focal loss.
The overall architecture of MARN is presented in Figure~\ref{fig:MARN}.
Firstly, we input word embeddings of clinical notes pretrained with the word2vec algorithm~\cite{mikolov2013distributed}.
Secondly, we adopt BiGRU as the feature extractor to extract textual representations from medical documents. 
Next, the RAM is plugged-in to improve the quality of document features learned by BiGRU and better handle noisy and lengthy clinical documents.
Then, we jointly train ICD and CCS coding branches by utilizing the multitask learning scheme to capture the associations among different medical codes. 
Finally, we utilize the focal loss~(FL) to alleviate the imbalanced class problem by redistributing the loss weight on high-frequency and low-frequency labels.

\subsection{Input Layer and Base Encoder}
Let $D$ be an clinical document consisting of $n$ tokens, $\{w_1, w_2, \cdots, w_n\}$.
The word embedding matrix is obtained by pretrained word2vec embeddings from each clinical document, denoted as $\mathbf{X}=[\mathbf{x}_1, \mathbf{x}_2, \cdots, \mathbf{x}_n ]^{\operatorname{T}}$, assembling to word a vector $\mathbf{x}_n$ whose word embedding size is $d_e$. 
We choose BiGRU as the backbone neural network to extract feature from clinical documents. 
Hidden states of BiGRU on token $x_i$~(where $i\in{1, 2, \ldots, n}$) are denoted as: 
\begin{align}
\overrightarrow{\mathbf{h}_i} &= \overrightarrow{\operatorname{GRU}}(\mathbf{x}_i, \overrightarrow{\mathbf{h}_{i-1}})\\
\overleftarrow{\mathbf{h}_i} &= \overleftarrow{\operatorname{GRU}}(\mathbf{x}_i, \overleftarrow{\mathbf{h}_{i+1}}),
\end{align}
where $\overrightarrow{\operatorname{GRU}}$ and $\overleftarrow{\operatorname{GRU}}$ denote forward and backward $\operatorname{GRU}$s, respectively. 
Bidirectional hidden states are obtained by horizontal concatenation of $\overrightarrow{\mathbf{h}_i}$ and $\overleftarrow{\mathbf{h}_i}$, which is presented as:
\begin{align}
\mathbf{h}_i = \operatorname{Concat}(\overrightarrow{\mathbf{h}_i}, \overleftarrow{\mathbf{h}_i})
\end{align}
The dimension of each directional GRU is set as $d_r$, so that the bidirectional hidden state $\mathbf{h}_i$ has dimension $\mathbb{R}^{2d_r}$.
The final hidden representation matrix is represented as $\mathbf{H} = [\mathbf{h}_1, \mathbf{h}_2, \dots, \mathbf{h}_n]^{\operatorname{T}} \in \mathbb{R}^{n \times 2d_r}$. 

\subsection{Recalibrated Aggregation Module}
We design \textbf{R}ecalibrated \textbf{A}ggregation \textbf{M}odule~(RAM) to provide the model with better representation learning from lengthy and noisy document features, which is inspired by the squeeze-and-excitation networks in the computer vision domain~\cite{hu2018squeeze}.
Three observations motivate to enhance the ability to handle the lengthy and noisy features.
1) When the document is very long, BiGRU also encounters a vanishing gradient problem, which may affect the model's performance \textcolor{black}{when stacking additional layers upon the recurrent layers}.
2) Lower-level semantic features extracted by BiGRU contain rich textural information with more document details, while higher-level document features capture abstract features with global receptive fields. 
The combination of multi-level features can benefit the representation learning from clinical documents.  
3) Higher-level semantic features provide contextual information for recalibrating noisy input document features. 
In the following, we will give the details about the calculation flow of the RAM and how RAM addresses the three aspects mentioned above. 
\begin{figure}[htbp]
\centering
\includegraphics[width=0.5\linewidth]{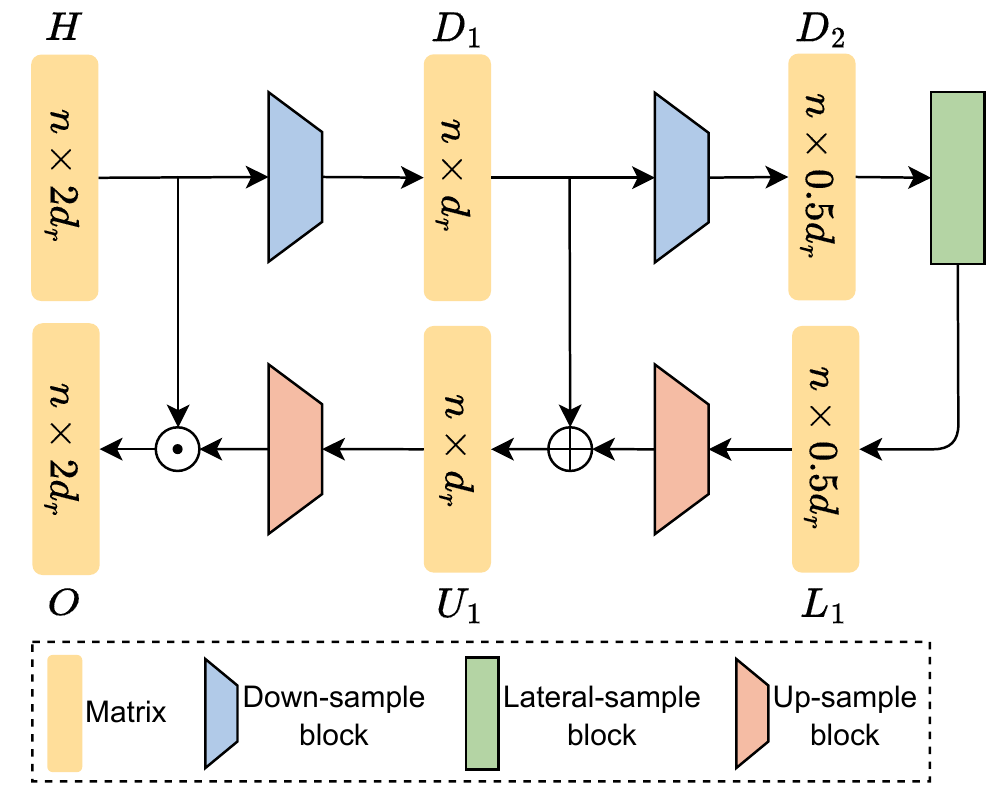}
\caption{The structure and feature flow of the RAM. ``$\odot$" denotes as element-wise multiplication. ``$\oplus$" represents the element-wise addition operation.}
\label{fig:RAM}
\end{figure}

\begin{figure}[!t]
\centering
\subfloat[\scriptsize{Down-sample Block}]{\includegraphics[width=0.2\textwidth]{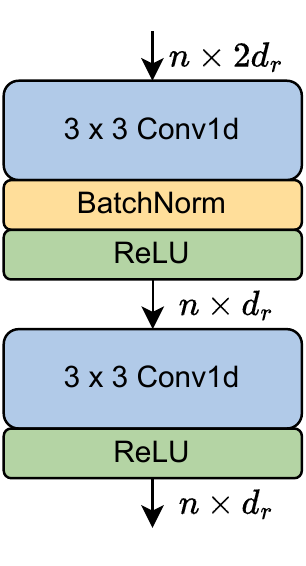}%
\label{fig:DB}}
\hfil
\subfloat[\scriptsize{Lateral-sample Block}]{\includegraphics[width=0.2\textwidth]{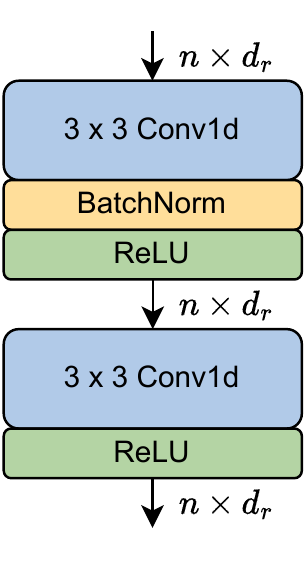}%
\label{fig:LB}}
\hfil
\subfloat[\scriptsize{Up-sample Block}]{\includegraphics[width=0.2\textwidth]{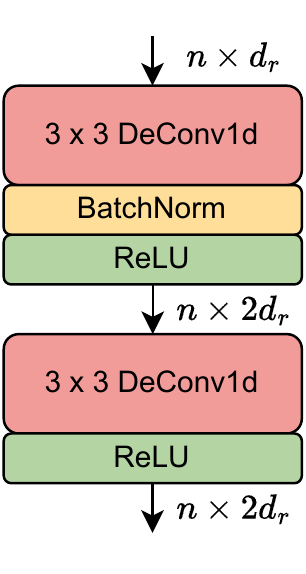}%
\label{fig:UB}}
\caption{Illustrations for the structure and dimension transformation of three basic blocks in RAM}
\label{fig:RAM_block}
\end{figure}

The structure and feature flow of RAM is illustrated in Figure~\ref{fig:RAM}.
The RAM consists of three kinds of blocks, i.e., ``Up-sample Block'', ``Lateral-sample Block'' and ``Down-sample Block'', whose structure and dimensionalities are demonstrated in Figure~\ref{fig:DB},~\ref{fig:LB},~\ref{fig:UB}, respectively.

We construct our basic blocks with stacked small convolution filters ~($kernel~ size = 3$) to have an appropriate receptive field~\cite{peng2017large}. 
To simplify the description, we regard the operations of ``Down-sample Block'',  ``Lateral-sample Block'' and ``Up-sample Block'' as $\mathcal{F_{DB}}(\cdot)$, $\mathcal{F_{LB}}(\cdot)$ and $\mathcal{F_{UB}}(\cdot)$.
The calculations of three blocks could be represented as following:

\begin{align}
\mathcal{F_{DB}}(\cdot) &= \mathcal{F}_{3 \times 3}^l (\operatorname{tanh}(\mathcal{F}_{3 \times 3}^d (\cdot)))\\
\mathcal{F_{LB}}(\cdot) &= \mathcal{F}_{3 \times 3}^l (\operatorname{tanh}(\mathcal{F}_{3 \times 3}^l (\cdot)))\\
\mathcal{F_{UB}}(\cdot) &= \mathcal{F}_{3 \times 3}^l (\operatorname{tanh}(\mathcal{F}_{3 \times 3}^u (\cdot)))
\end{align}
where $\mathcal{F}_{3 \times 3}^d$ is a $3 \times 3$ convolution layer followed by the BatchNorm operation, where the number of output channels is reduced to half of the input channels. 
The operation of $\mathcal{F}_{3 \times 3}^l$ is quite similar to $\mathcal{F}_{3 \times 3}^d$ except that $\mathcal{F}_{3 \times 3}^l$ retains the input channels' number. 
$\mathcal{F}_{3 \times 3}^u$ performs the deconvolution operation that doubles the input channels, and the output feature is passed through the Batch Normalization layer.

The RAM is divided into three stages: feature abstraction, feature smoothing, and feature aggregation.
\textcolor{black}{Firstly, the RAM utilizes cascaded convolutional blocks to get abstract features which have larger receptive fields by comparing with the input features of BiGRU.
Each convolutional layer incorporates $k$-grams~($k$ is the convolutional kernel size) into one-gram, so that large-scale n-grams' information can be embedded into one-gram if input features are passed through cascaded convolutional blocks.
In this way, the RAM has the better ability to handle lengthy clinical documents.
Secondly, the abstract features are processed to smooth the input features by a convolutional blocks whose input dimension sizes are the same as the output ones.
We denote this stage as the feature smoothing because of no variation on feature dimensions.
Thirdly, we use de-convolutional blocks to restore feature dimensions~(consistent receptive fields) \textcolor{black}{and avoid the feature inconsistence}\footnote{\textcolor{black}{Experimental results show that the de-convolutional up-sample block outperforms the simple linear projection across all scores by 0.05\%~1.73\%.}}, in order to facilitate the aggregation of abstract feature~(contain context-aware information) and lower-level feature (contain detailed document information) to generate richer document representations for recalibrating noisy input features.}
\textcolor{black}{When dealing with long clinical documents, the RAM module does not worsen the vanishing gradient problem caused by the BiGRU backbone because of identity paths in the convolutional blocks and the activation function.}  

\textbf{Feature Abstraction:} We leverage two Down-sample blocks to abstract the input document features $\mathbf{H}$.
The computation process is denoted as:
\begin{align}\label{eq:feature_abstract}
\mathbf{D_1} &= \mathcal{F_{DB}}(\mathbf{H}) \\
\mathbf{D_2} &= \mathcal{F_{DB}}(\mathbf{D_1}),
\end{align}
where the higher-level document features $\mathbf{D_1} \in \mathbb{R}^{n \times d_r}$ and $\mathbf{D_2} \in \mathbb{R}^{n \times 0.5d_r}$ are extracted as the input features for next two stages. 

\textbf{Feature Smoothing:} We use the Lateral-sample block to further extract high-level features and take the feature $\mathbf{D_2}$ as input, i.e., 
\begin{align}\label{eq:smooth_feature}
\mathbf{L_1} &= \mathcal{F_{LB}}(\mathbf{D_2}),
\end{align}
where the output feature of this stage is represented as $\mathbf{L_1} \in \mathbb{R}^{n \times 0.5d_r}$.

\textbf{Feature Aggregation:} Two Up-sample blocks are utilized to \textcolor{black}{recover} the dimension of the feature $\mathbf{L_1}$ \textcolor{black}{for keeping receptive field consistent when fusing the output features with original input features~$\mathbf{H}$}.
At the same time, we use two different fusion operations, element-wise addition and multiplication, to inject lower-level features $\mathbf{H}$ and $\mathbf{D_1}$ into features containing rich contextual information.
\textcolor{black}{The output features~$\mathbf{O}$ is followed by a $\operatorname{tanh}$ function to activate output features and a dropout layer is added after activated features to prevent overfitting.}
This conflation process is demonstrated as:
\begin{align}\label{eq:feature_aggregation}
\mathbf{U_1}&= \mathcal{F_{UB}}(\mathbf{L_1}) \oplus \mathbf{D_1},\\
\mathbf{O} &= \operatorname{tanh}(\mathcal{F_{UB}}(\mathbf{U_1}) \odot \mathbf{H}),
\end{align}
where $\mathbf{U_1} \in \mathbb{R}^{n \times d_r}$ is the fused feature and $\mathbf{O} \in \mathbb{R}^{n \times 2d_r}$ is the final output of RAM.

\subsection{Attention Classification Layers}
\begin{figure}[htbp]
\centering
\includegraphics[width=0.7\linewidth]{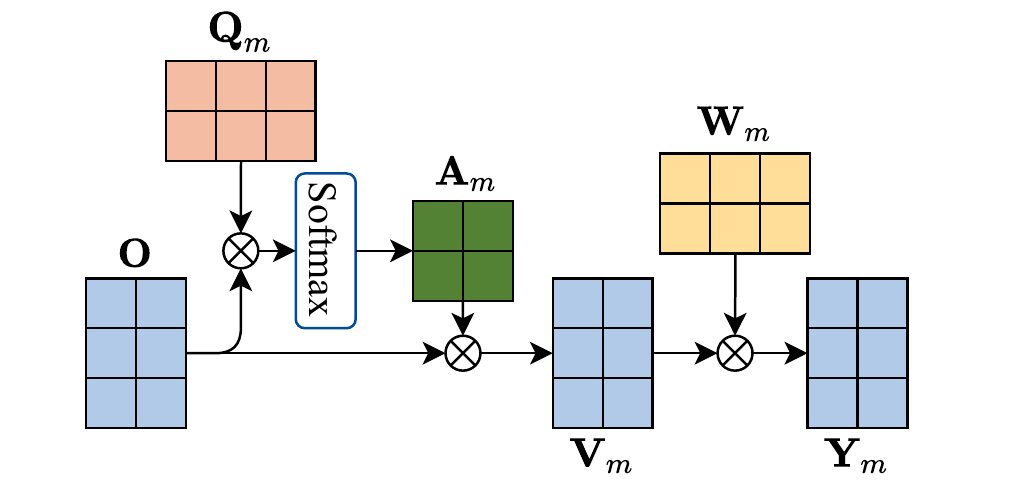}
\caption{An illustration of the label attention layer. ``$\otimes$" represents the matrix multiplication operation.}
\label{fig:Label_aware}
\end{figure}

The feature $\mathbf{O}$ output by the RAM is label-agnostic.
We deploy the label-aware attention mechanism to connect the label information of each medical code with different positions of the clinical document feature vector $\mathbf{O}$.
We set two medical coding branches, the ICD and CCS coding branches, with separate label-aware attention mechanisms shown in Figure~\ref{fig:Label_aware}.

To simplify the description of the two coding branches, we combine two subscripts of ICD and CCS codes into one subscript $m$ because the calculation are the same in two branches, where $m$ denotes either ICD or CCS code, and omit the bias item in the equations.
The attention score of each medical code is computed as:
\begin{equation}
\mathbf{A}_m = \operatorname{Softmax}(\mathbf{O}\mathbf{Q}_m),
\end{equation}
where $\mathbf{O}$ is the output feature from the RAM, $\mathbf{Q}_m \in \mathbb{R}^{2d_r \times d_m}$ denotes the trainable parameter matrix of the queries in the label-aware attention layer, and $d_m$ refers to the number of target medical codes.
The label attentive document features $\mathbf{V}_m \in \mathbb{R}^{d_m \times 2d_r}$ are generated by multiplying attention score matrix $\mathbf{A}_m$ with the calibrated feature $\mathbf{O}$ from RAM, i.e.,
\begin{equation}
\mathbf{V}_m = \mathbf{A}_m^{\operatorname{T}} \mathbf{O}.
\end{equation}
The label-aware attention mechanism can capture the code-related information and encode them into documents. 
Then, the label attentive features $\mathbf{V}_m$ are transformed into score vectors $\mathbf{Y}_m \in \mathbb{R}^{d_m \times 2d_r}$ of medical codes by using the fully-connected layer.
We pass the score vectors through the sum pooling operation followed by a Sigmoid activation function to generate probabilities $\mathbf{\hat{y}}_m$ for final medical code prediction, which are denoted as: 
\begin{align}
\mathbf{\mathbf{Y}}_m &= \mathbf{W}_m\mathbf{V}_m^{\operatorname{T}}\\
\mathbf{\hat{y}}_m &= \sigma(\operatorname{Pooling}(\mathbf{\mathbf{Y}}_m)),
\label{eq:output}
\end{align}
where $\mathbf{W}_m \in \mathbb{R}^{d_m \times 2d_r}$ are the learnable parameters of the fully-connected layer and $\sigma$ is the Sigmoid activate function. 

\subsection{Multitask Learning with Focal Loss}
\label{sec:MTL}
We perform multitask training to incorporate the two medical coding branches for the  ICD and CCS codes, respectively.
The probabilities of medical codes, $\mathbf{\hat{y}}_i$, produced by the label-aware classification layer are fed into the loss function separately.
The sparsity of codes poses a severe imbalanced class problem, which we alleviate using the focal loss when training the model~\cite{lin2017focal}. 
The focal loss of each medical coding branch $m$ can be written as: 
\begin{equation}
\mathcal{FL}_m = \sum_{i=1}^{d_m} [-y_i\alpha (1-\hat{y}_i)^{\gamma}\log(\hat{y}_i)
- (1 - y_i)(1-\alpha)\hat{y}_i^{\gamma}\log(1-\hat{y}_i) ],
\label{eq:fl}
\end{equation}
which is the sum of losses over all $d_m$ medical codes present in the $m$-th coding system. 
The parameter $\alpha$ represents the weighting factor that balances the loss for different classes ($y=1$ or $y=0$ depending on whether the code was present or not). 
We set $\alpha=0.999$, which places a strong emphasis on instances with a certain code present (i.e., $y=1$), to reflect the sparsity of the codes. 
In addition, the focal loss includes a modulating factor, $(1-\hat{y}_i)^{\gamma}$. 
If $\gamma>0$, the modulating factor places less emphasis on confident predictions and more emphasis on uncertain predictions, further making the model focus on the learning of codes that are difficult to classify, which often are the low-frequency codes. 
The loss weights of low- and high-frequency codes are adjusted by the confidences of prediction dynamically.
Note that for simplicity in (\ref{eq:fl}) we omitted the subscript $m$ that identifies the code branch in the ground truth labels $\mathbf{y}_m$ and prediction probabilities $\mathbf{\hat{y}}_m$. 

We treat the medical code prediction as a multitask problem, and consequently formulate the joint loss as: 
\begin{align}
\mathcal{FL}_M = \lambda_d \mathcal{FL}_d + \lambda_s \mathcal{FL}_s,
\end{align}
where $\mathcal{FL}_d$ and $\mathcal{FL}_s$ denote the focal losses for the ICD and CCS coding branches, and $\lambda_d$ and $\lambda_s$ are the loss weights of the ICD and CCS coding branches.

\section{Experiments}
\label{sec:exp}
In this section, we evaluate the effectiveness of our proposed model MARN on public real-world datasets. 
The source code is available at \url{https://github.com/VRCMF/MARN}. %

\subsection{Datasets}

\textbf{MIMIC-III (ICD codes):} 
The third version of Medical Information Mart for Intensive Care (MIMIC-III)\footnote{\url{https://mimic.physionet.org/gettingstarted/access/}} is a large, open-access dataset consisting of clinical data associated with above 40,000 inpatients in critical care units of the Beth Israel Deaconess Medical Center between 2001 and 2012~\cite{johnson2016mimic}. 
Following Mullenbach et al.~\cite{mullenbach2018explainable} and Li and Yu~\cite{li2020icd}, we use the discharge summaries as the input clinical documents.
Human experts annotate each summary document with corresponding diagnosis and procedure codes.
The first data set is the full data set with 8,921 unique ICD-9-CM codes in total.
MIMIC-III full codes data set has 52,722 discharge summaries, with 47,719, 1,631, and 3,372 documents for training, validation, and testing.
The second data set is the MIMIC-III top-50 codes data set, for which we divide all discharge summaries documents based on the patient IDs and generate the top 50 most frequent ICD codes.
The top-50 data set has 8,067 discharge summaries for training, and 1,574 and 1,730 documents for validation and testing, respectively. 
We refer two data set of ICD codes as \textbf{MIMIC-III-full (ICD codes)} and \textbf{MIMIC-III-50 (ICD codes)}.

\noindent \textbf{MIMIC-III (CCS codes):}
We leverage the ICD-CCS mapping scheme maintained by the HCUP~\footnote{\url{www.hcup-us.ahrq.gov/toolssoftware/ccs/ccs.jsp}} to project the ICD codes into a lower-dimensional CCS codes.
The top-50 ICD codes and full ICD codes~(8,921) are converted into top-38 CCS codes and full CCS codes~(295).
The MIMIC-III data sets of CCS codes share the same discharge summaries with the MIMIC-III ICD codes data set for training, validation, and testing.
We denote the full and the top-38 CCS code data sets as \textbf{MIMIC-III-full (CCS codes)} and \textbf{MIMIC-III-38 (CCS codes)} , respectively.

\subsection{Settings}

\noindent \textbf{Data Preprocessing:}
Following the previous works, we remove the non-alphabetic tokens, such as punctuation and numbers, from clinical documents. 
We transform all tokens into the lowercase format and change all tokens appearing in fewer than three notes into the `UNK' token. 
The medical word embeddings are established by the word2vec technique from all discharge summaries.
The dimension of word embedding $d_e$ is set as $100$, consistent with previous works.
We set the maximum length of each document as 4,000, with the exceeded part truncated. 

\noindent \textbf{Evaluation Metrics:}
We use the same evaluation metrics as previous works to validate the effectiveness of our proposed model on the data sets of two kinds of medical codes.
The evaluation metrics include macro-averaged and micro-averaged AUC-ROC~(area under the receiver operating characteristic curve),
macro-averaged and micro-averaged F1, precision at $k$ (dubbed as `P@$k$', where  $k \in \{8,15\}$).
P@$k$ is the precision score indicating the top-$k$ scored predictions in the ground truth labels. 

\noindent \textbf{Hyper-parameter Tuning:}
Our implementation details are as follows.
We train our model with the optimizer Adam~\cite{kingma2014adam} and set the learning rate to $0.001$.
The batch sizes of the top-n~($n \in \{38, 50\}$) and full code data are 16 and 64, respectively.
We apply the early stopping trick to exit the model training by monitoring the P@$k$ score, which avoids the model over-fitting. 
The training will stop if the P@$k$ score \textcolor{black}{on the validation set} does not improve in $10$ rounds.
We set the kernel size of each block in the RAM to 3. 
The dropout rate of the RAM is 0.2. 
For the multitask learning, we set ICD scaling factor $\lambda_d$ and CCS scaling factors $\lambda_s$ as 0.7 and 0.3, respectively. 
In the focal loss, the weighting factor $\alpha$ is 0.999, and the focusing parameter $\gamma$ is 2, tuned from 1 to 5. 
\textcolor{black}{Section~\ref{sec:exp_set} studies the detailed experimental settings about hyperparameters of the multitask learning and the focal loss.}

\subsection{Baselines}

\noindent \textbf{CNN}~\cite{mullenbach2018explainable}: The vanilla CNN model utilizes a max-pooling \textbf{C}onvolutional \textbf{N}eural \textbf{N}etwork~\cite{kim2014convolutional} to predict ICD codes.

\noindent \textbf{BiGRU}~\cite{mullenbach2018explainable}: This model uses a bidirectional recurrent architecture with gated recurrent units as the feature extractor for ICD coding.

\noindent \textbf{CAML}~\cite{mullenbach2018explainable}: \textbf{C}onvolutional \textbf{A}ttention network for \textbf{M}ulti-\textbf{L}abel classification (CAML) uses a convolutional neural network to extract the document features and the label-wise attention mechanism to enhance feature learning.

\noindent \textbf{DR-CAML}~\cite{mullenbach2018explainable}: \textbf{D}escription \textbf{R}egularized-\textbf{CAML}~(DR-CAML) is an extension model of the CAML, which incorporates textual descriptions of ICD codes to regularize the CAML model.

\noindent \textbf{MultiResCNN}~\cite{li2020icd}: \textbf{Mul}ti-Filter \textbf{Res}idual \textbf{C}onvolutional \textbf{N}eural \textbf{N}etwork (MultiResCNN) leverages a convolutional layer with multiple filters to capture various text patterns and adopts residual block to increase the receptive field on the model.

\noindent \textbf{LAAT}~\cite{vu2020label}: Vu et al. design the new \textbf{la}bel \textbf{at}tention model~(LAAT) by choosing bidirectional Long-Short Term Memory~(BiLSTM) as the feature extractor and deploying a label self-attention mechanism to learn label-specific vectors for ICD code predictions.

\noindent \textbf{JointLAAT}~\cite{vu2020label}: JointLAAT extends the LAAT by applying a hierarchical joint learning model to capture the hierarchical structure of ICD codes.

\noindent \textcolor{black}{\textbf{Fusion}~\cite{luo2021fusion}: Fusion utilizes compressed convolutional layer to encode clinical notes into informative local features, which are fused into output representations for ICD code predictions.}

\noindent \textcolor{black}{\textbf{MDBERT}~\cite{zhang2022hierarchical}: Medical Document BERT~(MDBERT) is a bottom-up hierarchical framework that combine features in word-level, sentence-level, and document-level to efficiently encode long documents. 
The hierarchical encoding model is first proposed by Yang et al.\cite{yang2016hierarchical} and applied to ICD coding by Dong et al.~\cite{dong2021explainable}.}

\subsection{Results}
\label{sec:results}

\noindent \textbf{MIMIC-III-50~(ICD code):}
Table~\ref{table:ICD50} shows the experimental results of baseline models and our proposed model on MIMIC-III-50~(ICD code) data set.
We observe that the MARN outperforms the other models~(CNN, BiGRU, CAML, DR-CAML, MultiResCNN \textcolor{black}{ and MT-RAM}) clearly across all evaluation metrics. 
The JointLAAT model uses a hierarchical joint learning mechanism to deal with the imbalanced class issue.
Compared with JointLAAT, our proposed model~(MARN) improves macro-AUC, micro-AUC, macro-F1, micro-F1, P@5 scores by 0.2\%, 0.1\%, 2.1\%, 0.2\% and 0.2\%, respectively.
The MARN has significant improvements, especially on the macro-F1 score, by 7.6\%, 10.6\%, 19.8\%, 7.6\%\textcolor{black}{, and 3.0\%} compared with MultiResCNN, DR-CAML, CAML, BiGRU, CNN\textcolor{black}{, and MT-RAM}. 
\textcolor{black}{Fusion improved all scores about 0.1\% $\sim$ 0.7\% compared with the MARN.}

In recent years, the pretrained language models with the transformer architecture such as BERT~\cite{devlin2019bert} have dominated many natural language processing tasks~\cite{lee2021fnet}. 
However, applying the BERT to medical coding tasks suffers from the limited document sequence~(512 tokens)~\cite{ji2021does} on the medical coding task.
\textcolor{black}{MDBERT, a BERT-based ICD coding framework, achieved competitive performance on the MIMIC-III-50~(ICD code) data set, while the MARN outperformed the MDBERT by 0.9\%, 1.1\%, 2.3\%, 2.6\%, and 1.9\% in macro-AUC, micro-AUC, macro-F1, micro-F1, and P@5 scores.}

\begin{table}[htbp]
\setlength\tabcolsep{4pt} %
\begin{center}
\begin{tabular*}{0.9\linewidth}{@{\extracolsep{\fill}} lrr|rr|r }
\toprule
\multicolumn{1}{l}{\multirow{2}{*}{Models}} & \multicolumn{2}{c}{AUC-ROC} & \multicolumn{2}{c}{F1} & \multirow{2}{2em}{P@5}   \\
& \multicolumn{1}{c}{Macro}       & \multicolumn{1}{c|}{Micro}      & \multicolumn{1}{c}{Macro}      & \multicolumn{1}{c|}{Micro}      &      \\ 
\midrule
\multicolumn{1}{l}{\textbf{CNN}} & 87.6 & 90.7 & 57.6 & 62.5 & 62.0\\
\multicolumn{1}{l}{\textbf{BiGRU}} & 82.8 & 86.8 & 48.4 & 54.9 & 59.1\\
\multicolumn{1}{l}{\textbf{CAML}} & 87.5& 90.9& 53.2& 61.4& 60.9\\ 
\multicolumn{1}{l}{\textbf{DR-CAML}}& 88.4  & 91.6 & 57.6 & 63.3& 61.8\\ 
\multicolumn{1}{l}{\textbf{MultiResCNN}}& 89.9 & 92.8& 60.6& 67.0& 64.1\\
\multicolumn{1}{l}{\textbf{\textcolor{black}{MT-RAM}}}& \textcolor{black}{92.1} & \textcolor{black}{94.3} & \textcolor{black}{65.2} & \textcolor{black}{70.7} & \textcolor{black}{66.4}\\
\multicolumn{1}{l}{\textbf{LAAT}}& 92.5 & 94.6 & 66.6& 71.5& {67.5}\\
\multicolumn{1}{l}{\textbf{JointLAAT}}& 92.5 & 94.6& 66.1& 71.6& 67.1\\
\multicolumn{1}{l}{\textbf{\textcolor{black}{Fusion}}}& \textbf{\textcolor{black}{93.1}} & \textbf{\textcolor{black}{95.0}}& \textbf{\textcolor{black}{68.3}}& \textbf{\textcolor{black}{72.5}}& \textbf{\textcolor{black}{67.9}}\\
\multicolumn{1}{l}{\textbf{\textcolor{black}{MDBERT}}}& \textcolor{black}{91.8} & \textcolor{black}{93.6}& \textcolor{black}{65.9}& \textcolor{black}{69.2}& \textcolor{black}{65.4}\\
\hline
\multicolumn{1}{l}{{MARN}(ours)} & {92.7} & {94.7}  & {68.2} & {71.8}& 67.3 \\
\bottomrule
\end{tabular*}
\end{center}
\captionsetup{justification=centering}
\caption{MIMIC-III-50~(ICD code) data set results~(in \%).}
\label{table:ICD50}
\end{table}

\noindent \textbf{MIMIC-III-full~(ICD code):}
Table~\ref{table:ICDfull} shows the results of the MARN and other strong baseline models.
The MARN performs better on macro-F1, micro-F1, P@8, and P@15 scores than other baseline models.
When compared with the state-of-the-art model~(JointLAAT), our proposed model has improved the scores of macro-F1, micro-F1, P@8, P@15 by 0.9\%, 0.9\%  1.9\%, and 1.2\%, respectively. 
Compared to convolution-based models including CNN, CAML, DR-CAML, MultiResCNN, \textcolor{black}{and Fusion,} the MARN vastly increases the micro-F1 score by 16.5\%, 4.5\%, 5.5\%, 3.2\%, \textcolor{black}{3.0\%} respectively.
\textcolor{black}{The MARN improved the macro-F1, micro-F1, p@8, and p@15 scores by 1.5\%, 2.9\%, 2.7\%, 2.5\%, comparing with the BERT-based MDBERT model.}

\begin{table}[htbp]
\setlength\tabcolsep{4pt} %
\begin{center}
\begin{tabular*}{0.9\linewidth}{@{\extracolsep{\fill}} lrr|rr|rr }
\toprule
\multicolumn{1}{l}{\multirow{2}{*}{Models}} & \multicolumn{2}{c}{AUC-ROC} & \multicolumn{2}{c}{F1} & \multicolumn{2}{c}{P@k}   \\
& \multicolumn{1}{c}{Macro}       & \multicolumn{1}{c|}{Micro}      & \multicolumn{1}{c}{Macro}      & \multicolumn{1}{c|}{Micro}      &   
\multicolumn{1}{c}{8}      & \multicolumn{1}{c}{15}     \\ 
\midrule
\multicolumn{1}{l}{\textbf{CNN}} & 80.6 & 96.9 & 4.2 & 41.9 & 58.1& 44.3\\
\multicolumn{1}{l}{\textbf{BiGRU}} & 82.2 & 97.1 & 3.8 & 41.7 & 58.5& 44.5\\
\multicolumn{1}{l}{\textbf{CAML}} & 89.5 & 98.6 & 8.8 & 53.9 & 70.9& 56.1\\
\multicolumn{1}{l}{\textbf{DR-CAML}}& 89.7  & 98.5 & 8.6 & 52.9& 69.0& 54.8\\ 
\multicolumn{1}{l}{\textbf{MultiResCNN}}& 91.0 & 98.6& 8.5& 55.2& 73.4& 58.4\\
\multicolumn{1}{l}{\textbf{LAAT}}& 91.9 & 98.8& 9.9 & 57.5& 73.8& 59.1\\
\multicolumn{1}{l}{\textbf{JointLAAT}}& {92.1} & 98.8& 10.7& 57.5& 73.5& 59.0\\
\multicolumn{1}{l}{\textbf{\textcolor{black}{Fusion}}}& {\textcolor{black}{91.5}} & \textcolor{black}{98.7}& \textcolor{black}{8.3}& \textcolor{black}{55.4}& \textcolor{black}{73.6}& - \\
\multicolumn{1}{l}{\textbf{\textcolor{black}{MDBERT}}}& \textbf{\textcolor{black}{92.5}} & \textbf{\textcolor{black}{98.9}}& \textcolor{black}{10.1}& \textcolor{black}{55.5}& \textcolor{black}{72.7}& \textcolor{black}{57.7} \\
\hline
\multicolumn{1}{l}{\textbf{MARN}(ours)} & 91.3 & {98.8}  & \textbf{11.6} & \textbf{58.4}& \textbf{75.4} & \textbf{60.2} \\
\bottomrule
\end{tabular*}
\end{center}
\captionsetup{justification=centering}
\caption{MIMIC-III-full~(ICD code) data set results~(in \%).}
\label{table:ICDfull}
\end{table}

\noindent \textbf{MIMIC-III-50~(CCS code):}
We validate the BiGRU, CAML, DR-CAML, and the MultiResCNN on the MIMIC-III-50~(CCS code) dataset and show the evaluation results in Table~\ref{table:CCS50}.
The MARN outperforms all baseline models by large margins across all evaluation metrics. 
Significantly, MARN improves the macro-F1 and micro-F1 scores by 8.3\% and 6.2\% compared with the MultiResCNN.
Our model also outperforms BiGRU, CAML, DR-CAML on macro-F1 and micro-F1 scores with 10\% $\sim$ 14\% and 8\% $\sim$ 13\%, respectively. 

\begin{table}[htbp]
\setlength\tabcolsep{4pt} %
\begin{center}
\begin{tabular*}{0.8\linewidth}{@{\extracolsep{\fill}} lrr|rr|r }
\toprule
\multicolumn{1}{l}{\multirow{2}{*}{Models}} & \multicolumn{2}{c}{AUC-ROC} & \multicolumn{2}{c}{F1} & \multirow{2}{2em}{P@5}   \\
& \multicolumn{1}{c}{Macro}       & \multicolumn{1}{c|}{Micro}      & \multicolumn{1}{c}{Macro}      & \multicolumn{1}{c|}{Micro}      &      \\ 
\midrule
\multicolumn{1}{l}{\textbf{BiGRU}} & 87.6 & 90.7 & 57.6 & 62.5 & 62.0\\
\multicolumn{1}{l}{\textbf{CAML}} & 89.2& 92.2& 60.9& 67.5& 64.5\\ 
\multicolumn{1}{l}{\textbf{DR-CAML}}& 87.5  & 90.5 & 59.3 & 65.6& 62.6\\ 
\multicolumn{1}{l}{\textbf{MultiResCNN}}& 89.2 & 92.4& 62.9& 68.8& 64.6\\
\hline
\multicolumn{1}{l}{\textbf{MARN}(ours)} & \textbf{92.8} & \textbf{95.0}  & \textbf{71.2} & \textbf{75.0}& \textbf{69.0} \\
\bottomrule
\end{tabular*}
\end{center}
\captionsetup{justification=centering}
\caption{MIMIC-III-50 results (CCS code) data set results~(in \%).}
\label{table:CCS50}
\end{table}

\noindent \textbf{MIMIC-III-full~(CCS code):}
We also evaluate the same baseline models on the MIMIC-III full~(CCS code) and compare our model to verify the effectiveness of the MARN.
Table~\ref{table:CCSfull} shows our model improves the macro-F1 score by 7.8\%, compared with the MultiResCNN model.
Our model also promotes other evaluation scores, with macro-AUC, micro-AUC, micro-F1, P@8, and P@15 increased by 3.1\%, 0.9\%, 3.2\%, 2.5\%, and 2.4\%, respectively. 

\begin{table}[htbp]
\setlength\tabcolsep{4pt} %
\begin{center}
\begin{tabular*}{0.9\linewidth}{@{\extracolsep{\fill}} lrr|rr|rr }
\toprule
\multicolumn{1}{l}{\multirow{2}{*}{Models}} & \multicolumn{2}{c}{AUC-ROC} & \multicolumn{2}{c}{F1} & \multicolumn{2}{c}{P@k}   \\
& \multicolumn{1}{c}{Macro}       & \multicolumn{1}{c|}{Micro}      & \multicolumn{1}{c}{Macro}      & \multicolumn{1}{c|}{Micro}      &   
\multicolumn{1}{c}{8}      & \multicolumn{1}{c}{15}     \\ 
\midrule
\multicolumn{1}{l}{\textbf{BiGRU}}& 91.2 & 96.4& 50.1& 68.4& 81.1& 64.0\\ %
\multicolumn{1}{l}{\textbf{CAML}} & 88.8 & 96.1& 44.4& 66.5& 80.5& 63.6\\ %
\multicolumn{1}{l}{\textbf{DR-CAML}}& 85.7 & 95.5& 41.3& 66.0& 78.9& 62.5\\ %
\multicolumn{1}{l}{\textbf{MultiResCNN}}& 90.6 & 96.5& 50.8& 69.0& 81.8& 64.8\\ %
\hline
\multicolumn{1}{l}{\textbf{MARN}(ours)} & \textbf{93.9} & \textbf{97.4}  & \textbf{58.6} & \textbf{72.2}& \textbf{84.3} & \textbf{67.2} \\
\bottomrule
\end{tabular*}
\end{center}
\captionsetup{justification=centering}
\caption{MIMIC-III-full~(CCS code) data set results~(in \%).}
\label{table:CCSfull}
\end{table}

\subsection{Detailed Analysis of MARN}
\label{sec:analysis}

This section studies the properties of the proposed MARN model through several research questions.

\subsubsection*{\textbf{How does each component of MARN affect the prediction? }}
We conduct experiments to validate the effectiveness of each component of the MARN on the MIMIC-III-50~(ICD code) and MIMIC-III-full~(ICD code) datasets, with the following specific building components considered:
\begin{itemize}
\item \textbf{M}ulti-\textbf{t}ask \textbf{L}earning scheme~(MTL)
\item \textbf{R}ecalibrated \textbf{A}ggregation \textbf{M}odule~(RAM)
\item \textbf{F}ocal \textbf{L}oss~(FL)
\end{itemize}
From Table~\ref{table:general_50} and Table~\ref{table:general_full}, we can observe that all components contribute to the performance improvement, and they are complementary to each other. 
The multitask learning scheme has a higher performance gain on the MIMIC-III-50~(ICD code) data set than the RAM, while the MIMIC-III-full~(ICD code) data set shows the opposite situation.
The model optimized with the focal loss outperforms the one with BCE loss. 

\begin{table*}[htbp]
\setlength\tabcolsep{4pt} %
\begin{center}
\begin{tabular*}{0.9\textwidth}{@{\extracolsep{\fill}} lrr|rr|r}
\toprule
\multicolumn{1}{l}{\multirow{2}{*}{Models}} & 
\multicolumn{2}{c}{AUC-ROC} & 
\multicolumn{2}{c}{F1} & 
\multirow{2}{2em}{P@5}\\
& \multicolumn{1}{c}{Macro}       & \multicolumn{1}{c|}{Micro}      & \multicolumn{1}{c}{Macro}      & \multicolumn{1}{c|}{Micro}      &
  \\ 

\midrule
\multicolumn{1}{l}{\textbf{MARN}} & \textbf{92.7} & \textbf{94.7} & \textbf{68.2} & \textbf{71.8} & \textbf{67.3}\\
\hline
\multicolumn{1}{l}{~w/o~$\operatorname{MTL}$}& 91.9 & 94.0 & 64.4 & 69.4 & 66.1\\
\multicolumn{1}{l}{~w/o~$\operatorname{MTL+FL}$}& 91.7& 93.4& 62.4& 68.1& 64.7\\ %
\multicolumn{1}{l}{~w/o~$\operatorname{RAM}$}& 92.3 & 94.3& 64.8& 69.9& 66.5\\ %
\multicolumn{1}{l}{~w/o~$\operatorname{RAM+FL}$}& 91.8 & 94.1& 64.6& 69.9& 66.2\\  %
\multicolumn{1}{l}{~w/o~$\operatorname{MTL+RAM+FL}$}& 91.2 & 93.4 & 59.2 & 67.2 & 65.5\\ %
\bottomrule
\end{tabular*}
\end{center}
\captionsetup{justification=centering}
\caption{Ablation results~(in \%) of MIMIC-III-50~(ICD code)}
\label{table:general_50}
\end{table*}

\begin{table*}[htbp]
\setlength\tabcolsep{4pt} %
\begin{center}
\begin{tabular*}{0.9\textwidth}{@{\extracolsep{\fill}} lrr|rr|rr}
\toprule
\multicolumn{1}{l}{\multirow{2}{*}{Models}} & 
\multicolumn{2}{c}{AUC-ROC}& 
\multicolumn{2}{c}{F1} & 
\multicolumn{2}{c}{P@k} \\
& \multicolumn{1}{c}{Macro}       & \multicolumn{1}{c|}{Micro}      & \multicolumn{1}{c}{Macro}      & \multicolumn{1}{c|}{Micro}      &   
\multicolumn{1}{c}{8}      & \multicolumn{1}{c}{15}     \\ 

\midrule
\multicolumn{1}{l}{\textbf{MARN}}& \textbf{91.3}& \textbf{98.8} & \textbf{11.6} & \textbf{58.4} & \textbf{75.4} & \textbf{60.2}\\
\hline
\multicolumn{1}{l}{~w/o~$\operatorname{MTL}$}& 89.9 & 98.6 & 10.5 & 57.1 & 73.5& 58.5\\
\multicolumn{1}{l}{~w/o~$\operatorname{MTL+FL}$}& 89.1 &98.4 &9.1 & 55.6 & 72.9 & 58.0\\ %
\multicolumn{1}{l}{~w/o~$\operatorname{RAM}$}& 90.4& 98.6 & 10.3 & 56.0 & 72.6 & 57.9\\ %
\multicolumn{1}{l}{~w/o~$\operatorname{RAM+FL}$}& 88.8& 98.2 & 7.4 & 50.7 & 69.5 & 54.8\\  %
\multicolumn{1}{l}{~w/o~$\operatorname{MTL+RAM+FL}$}& 88.9 & 98.3 & 6.8 & 51.5 & 69.9& 54.7\\ %
\bottomrule
\end{tabular*}
\end{center}
\captionsetup{justification=centering}
\caption{Ablation results~(in \%) of MIMIC-III-full~(ICD code).}
\label{table:general_full}
\end{table*}

\subsubsection*{\textbf{How compatible are the building blocks with different base models? }}
\begin{figure}[htbp]
\centering
\includegraphics[width=0.5\linewidth]{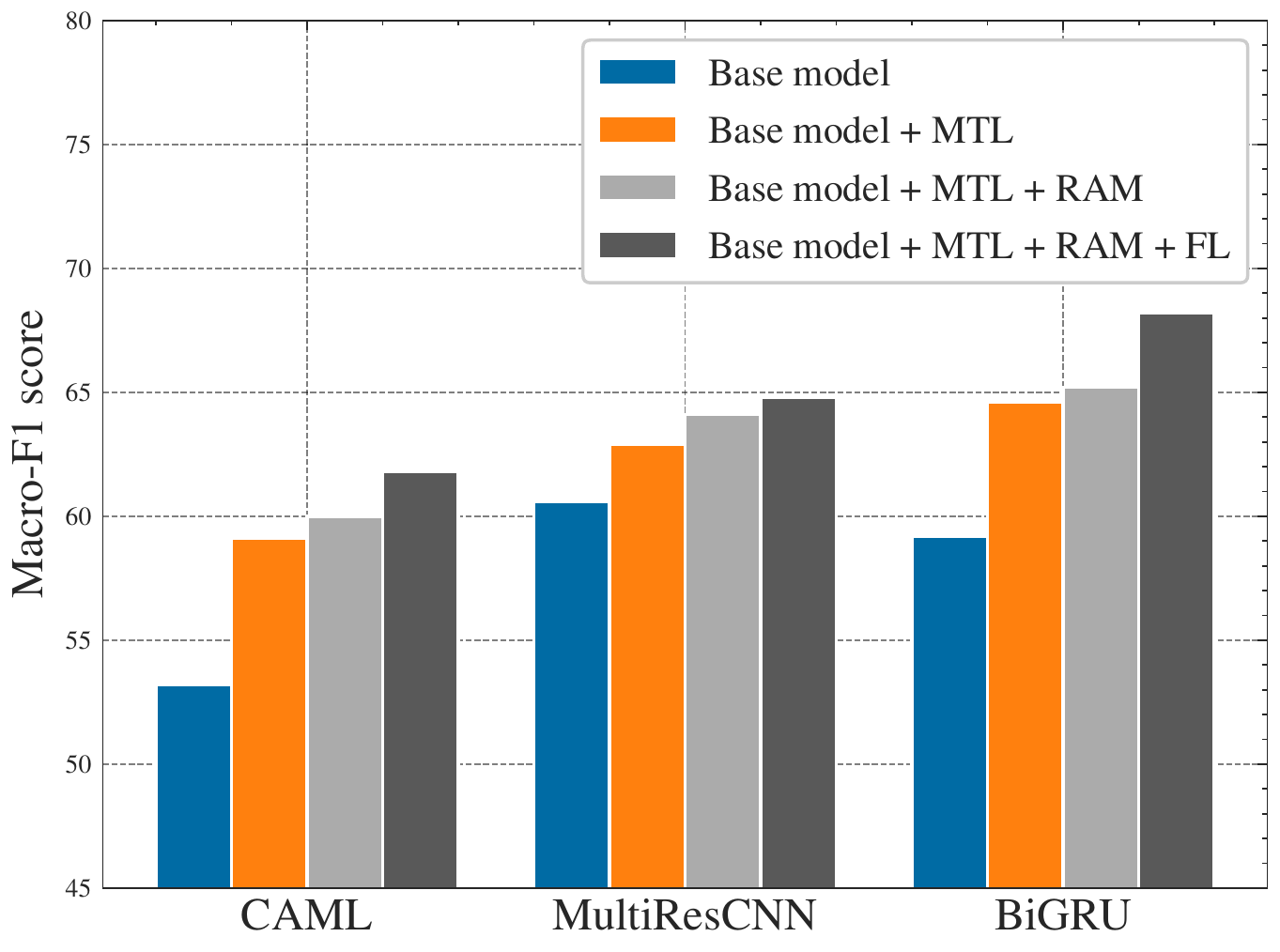}
\caption{Macro-F1 scores~(in \%) of different base models including the different building blocks of MARN}
\label{fig:coupling}
\end{figure}

We choose two convolution-based models~(i.e., CAML, MultiResCNN) and an RNN model~(BiGRU) to explore the compatibility of different building modules. 
We term these models as base models. 
Figure~\ref{fig:coupling} shows that the MTL, RAM, and FL can improve other base models' performance.
Significantly, the BiGRU model optimized with focal loss gains better performance than the CAML and MultiResCNN. 

\subsubsection*{\textbf{Can multitask learning connect different medical coding systems?}}

\begin{figure}[!htbp]
\centering
\includegraphics[width=0.6\linewidth]{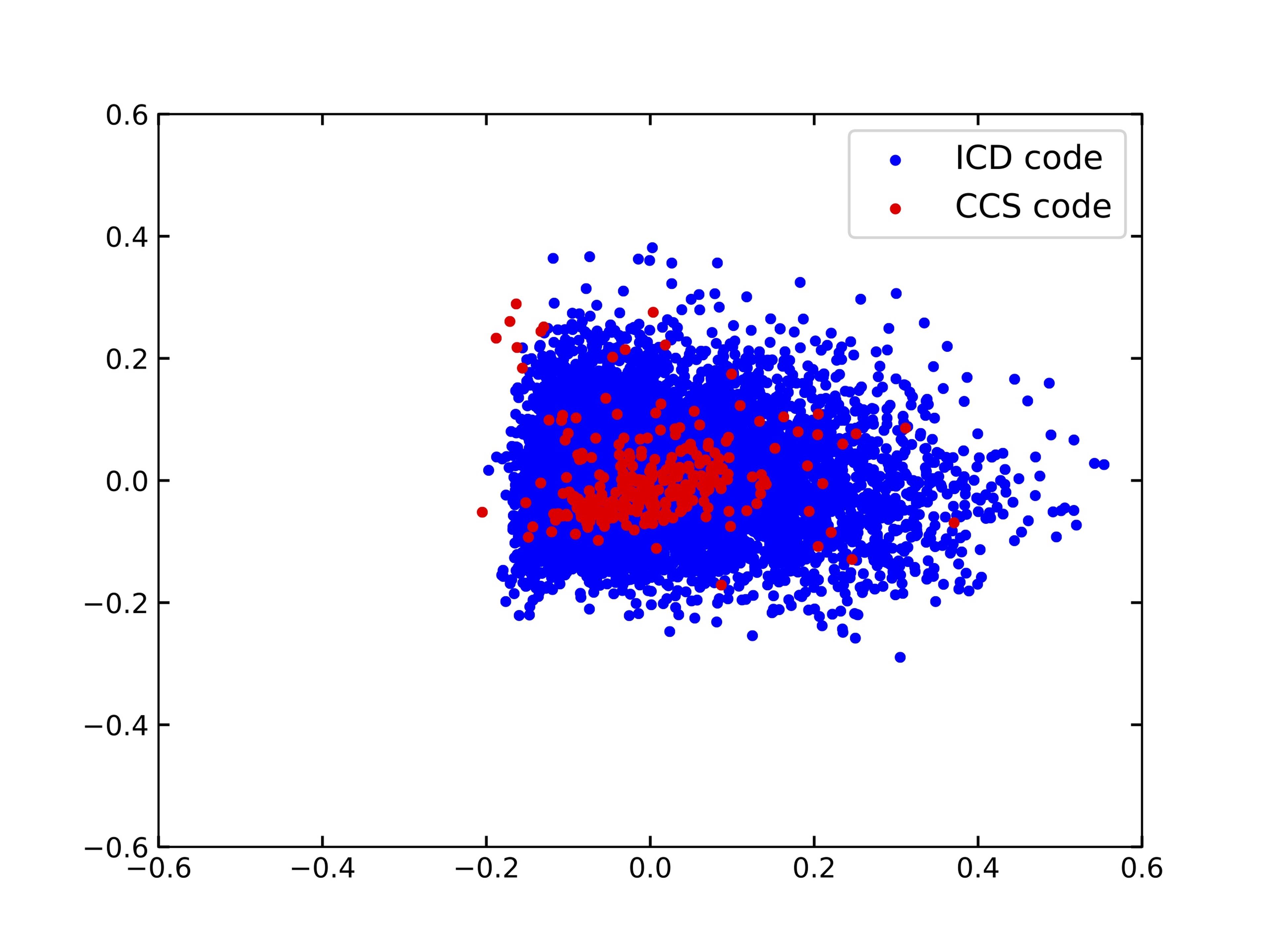}
\caption{The first two principal components of ICD and CCS code embeddings.}
\label{fig:full_dist}
\end{figure}

\begin{figure*}[!htbp]
\centering
\subfloat[\scriptsize{CCS code: 3}]{\includegraphics[width=0.5\textwidth]{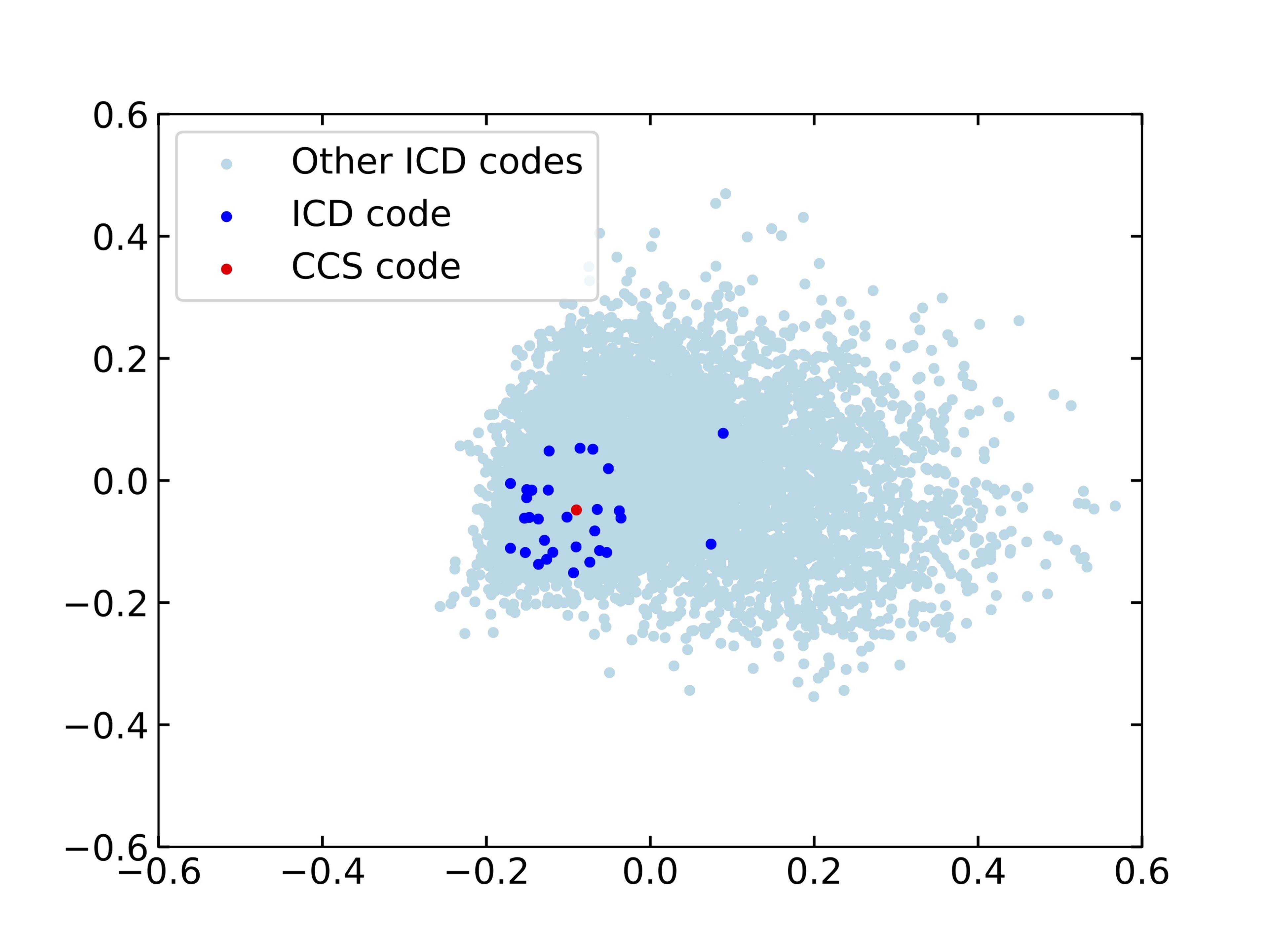}%
\label{fig:3}}
\hfil
\subfloat[\scriptsize{CCS code: 11}]{\includegraphics[width=0.5\textwidth]{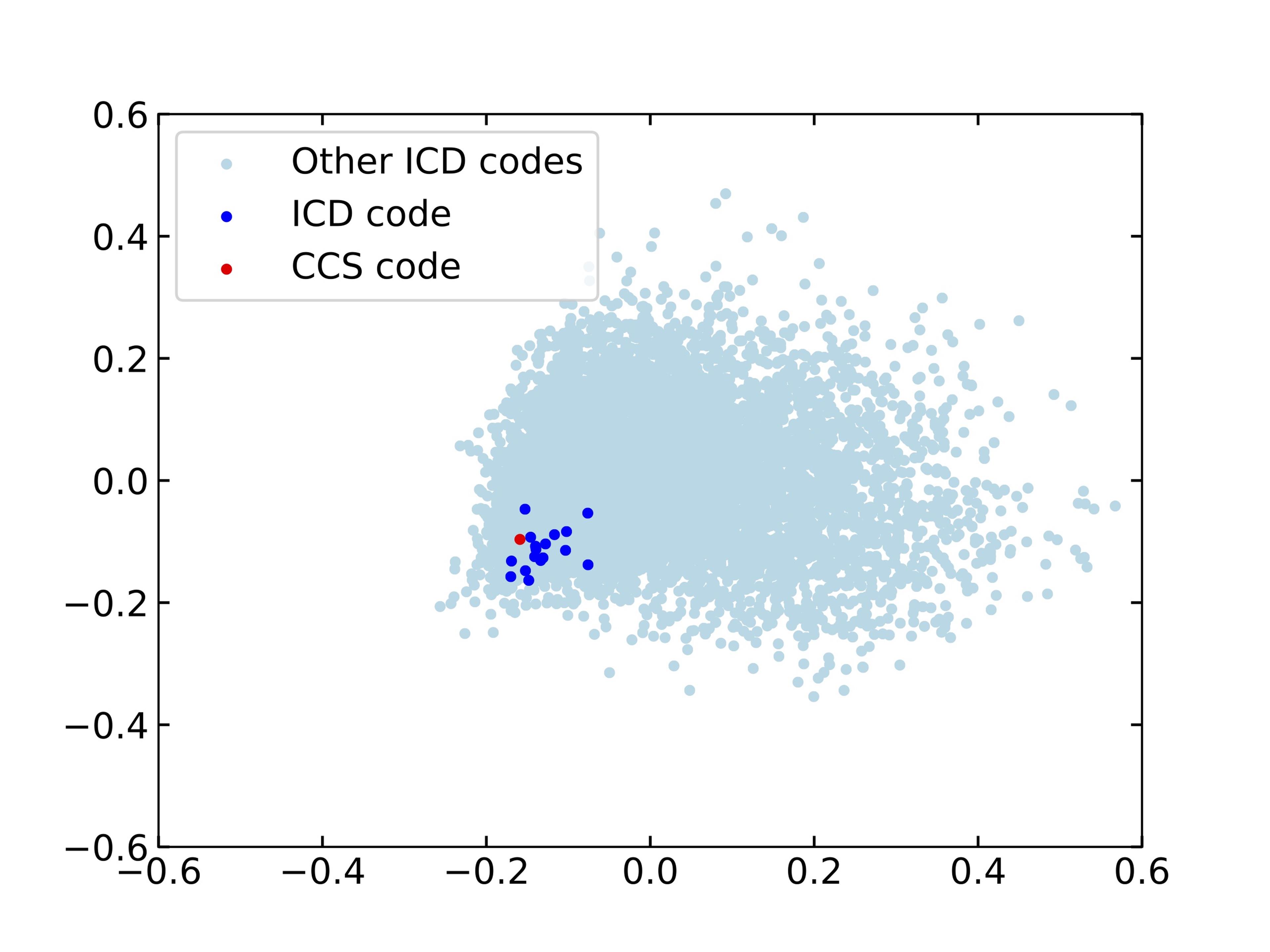}%
\label{fig:11}}
\\
\subfloat[\scriptsize{CCS code: 195}]{\includegraphics[width=0.5\textwidth]{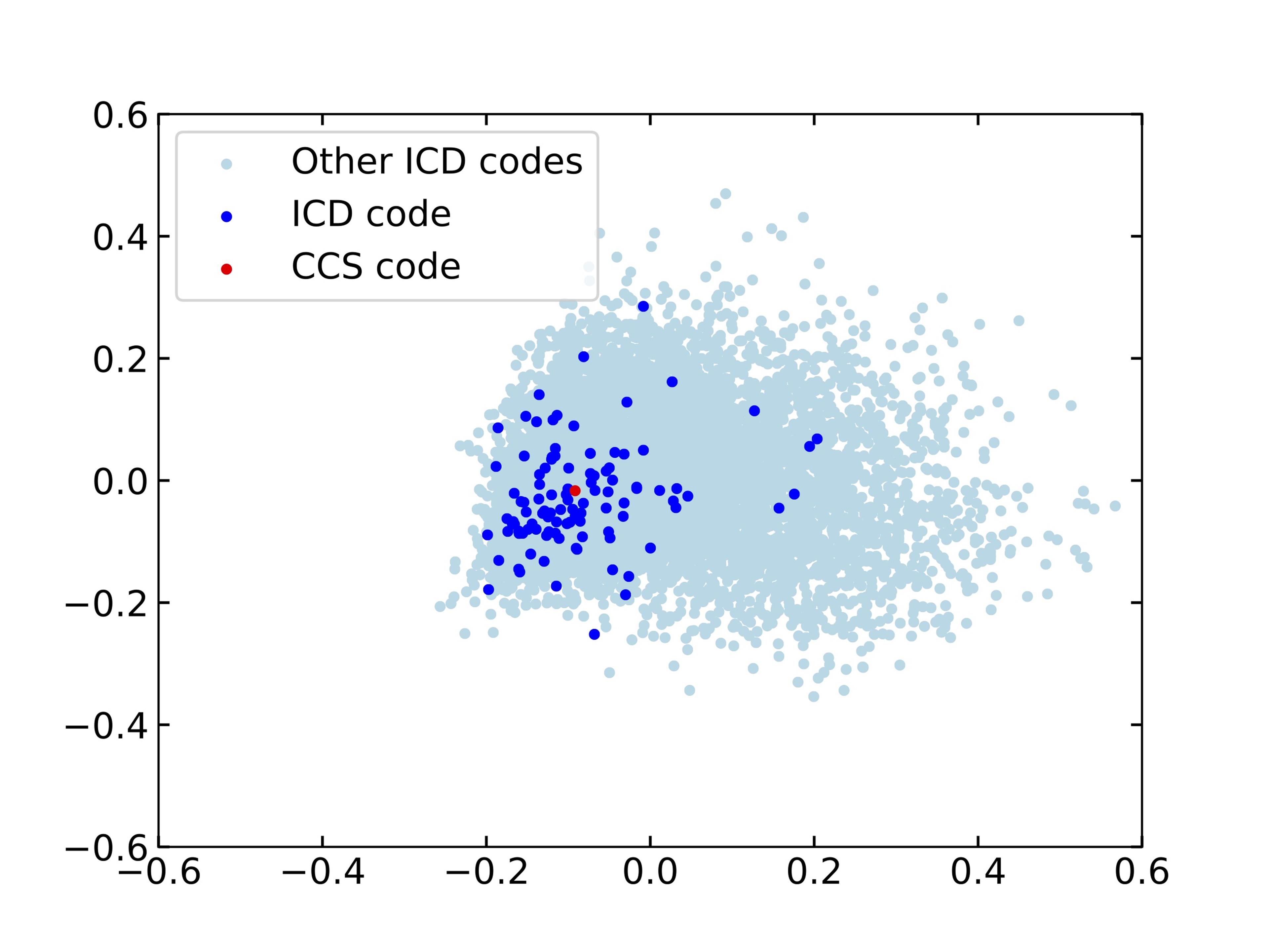}%
\label{fig:195}}
\hfil
\subfloat[\scriptsize{CCS code: 223}]{\includegraphics[width=0.5\textwidth]{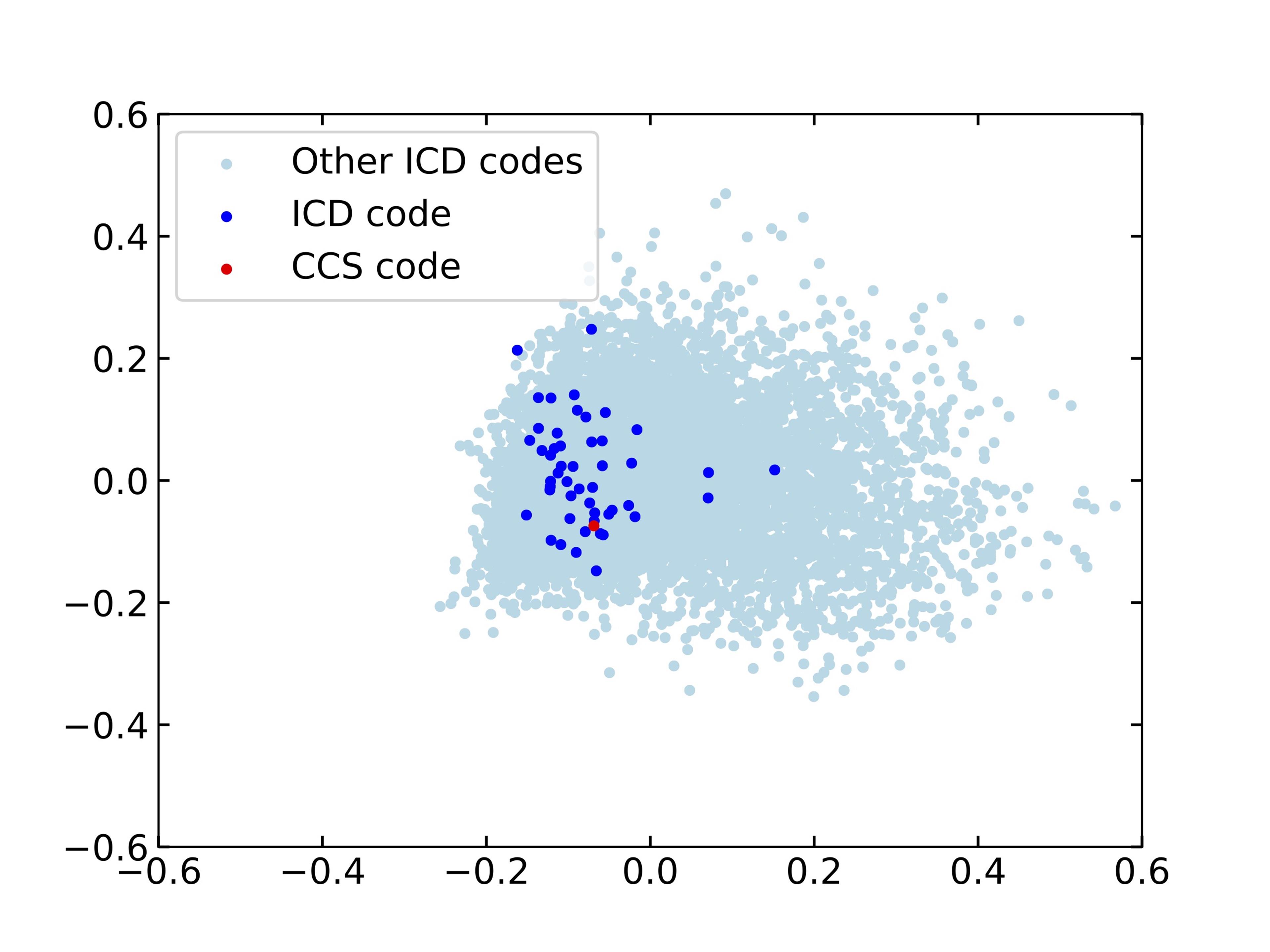}%
\label{fig:223}}
\caption{The embeddings of representative significant CCS codes and their corresponding ICD codes. We see that the relevant ICD codes are clustered around the respective significant CCS code, reflecting the ability of MARN to learn representations that capture informative relationships between the codes.}
\label{fig:MTL_plot}
\end{figure*}

We leverage Principal Component Analysis~(PCA) to reduce the dimension of document features and plot the embeddings of ICD codes and CCS codes in the resulting two-dimensional space.
Figure~\ref{fig:full_dist} demonstrates that the cloud of ICD codes surrounds the cluster of CCS codes, reflecting the mapping from ICD to CCS codes.
We also notice that ICD codes associated with a particular CCS code are often clustered around the respective CCS code, as shown for several representative CCS codes and their corresponding ICD codes in Fig.~\ref{fig:MTL_plot}.
To study this phenomenon in detail, we define a circular region around each CCS code using 10\% of the longest distance between two ICD codes as the radius. 
We then calculate the number of significant CCS codes, defined as CCS codes, such that their respective regions contain significantly more \textit{relevant} ICD codes than expected by chance, where relevant ICD codes are those that are known to map to the CCS code. 
We approximate the distribution of the number of relevant ICD codes within a region using the binomial distribution: 
\begin{align}
P(k) = Binomial(k|n,p)%
\end{align}
where $n$ is the total number of ICD codes in the region, and $p$ is the overall proportion of relevant ICD codes for the respective CCS code. 
For each CCS code, we define a threshold $T$~(the number of relevant ICD codes in the CCS code region), such that $P(k>T)<0.1$, and if the observed number of relevant codes within a region exceeds the threshold, we consider the CCS code as significant.
As a result, there are 157 significant CCS codes among the total of 295 CCS codes. 
Hence, we conclude that the MTL can establish informative connections between related ICD codes to benefit medical code prediction.

\subsubsection*{\textbf{Does the model optimized with focal loss balance the learning between low- and high-frequency codes?}}

\begin{figure}[htbp]
\centering
\includegraphics[width=0.6\linewidth]{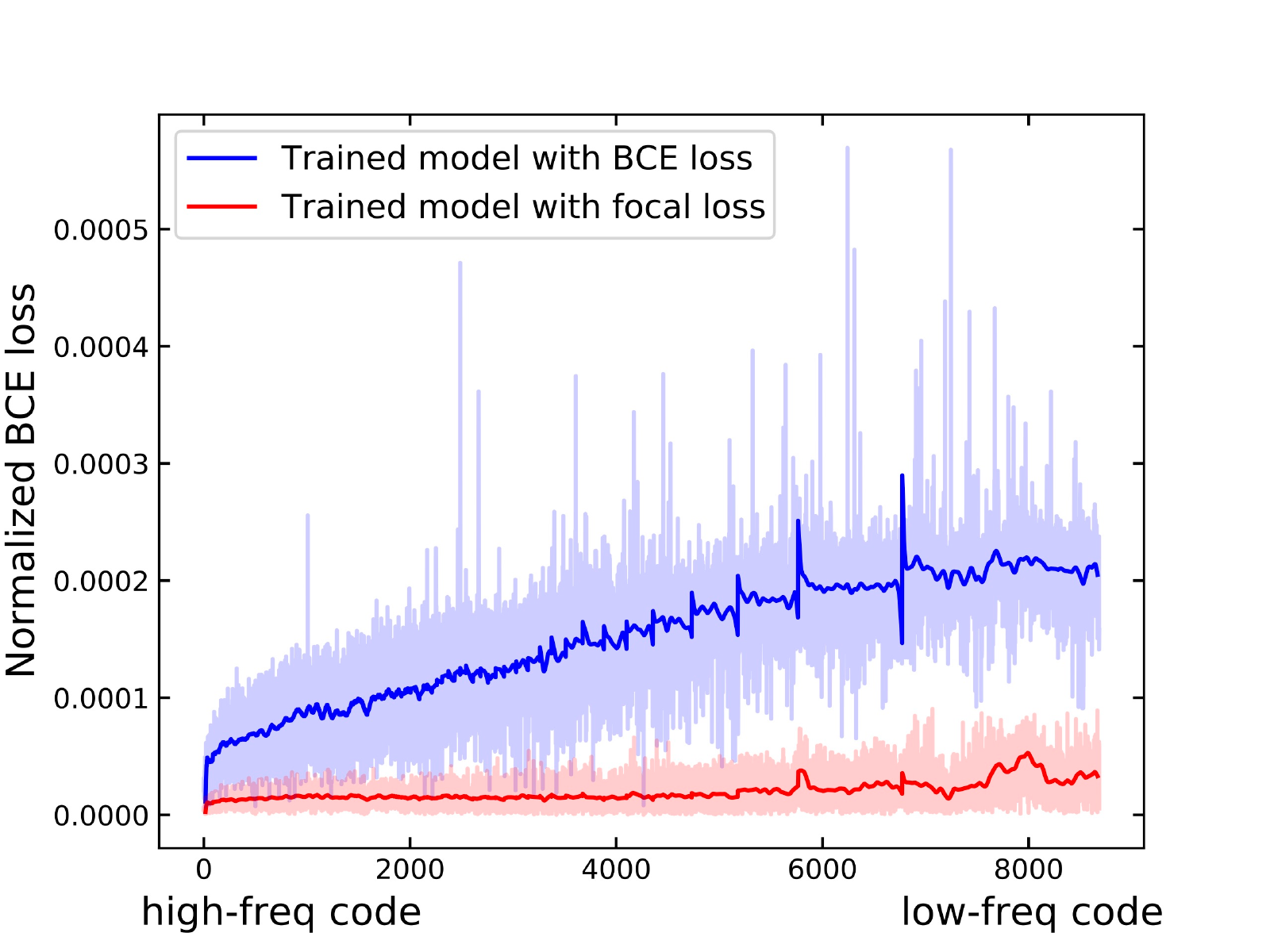}
\caption{Normalized binary cross entropy loss of each ICD code, with x-axis sorted by code frequency. The high-frequency codes are on the left, the low-frequency codes on the right. }
\label{fig:soft_fl}
\end{figure}

We plot the normalized loss value of each code to explore whether the focal loss can balance the loss of high- and low-frequency ICD codes.
Firstly, we train the model with different loss functions, take a forward pass of the trained model, and calculate a unified loss function, i.e., binary cross entropy~(BCE) loss, for a fair comparison.
Then, we normalize the loss of each ICD code by dividing the BCE loss by the frequency of the code and the total number of documents. 
Fig~\ref{fig:soft_fl} shows the normalized loss curve of models trained by the BCE loss and focal loss.
We see that optimizing the models with focal loss improves results overall especially for low-frequency codes. 
The normalized loss of the model optimized by focal loss is balanced compared with the model optimized by BCE loss.
Thus, we can conclude that the model optimized with BCE loss can not balance the learning of high- and low-frequency codes. 
In contrast, the model optimized with focal loss can effectively handle the imbalanced class problem in this study.

\subsection{Hyperparameter Studies}
\label{sec:exp_set}
\textcolor{black}{
We study two hyperparameters of the multitask learning scheme and the focal loss in this section.
}

\begin{figure}[htbp]
\centering
\begin{subfigure}[]{0.45\textwidth}
    \includegraphics[width=\textwidth]{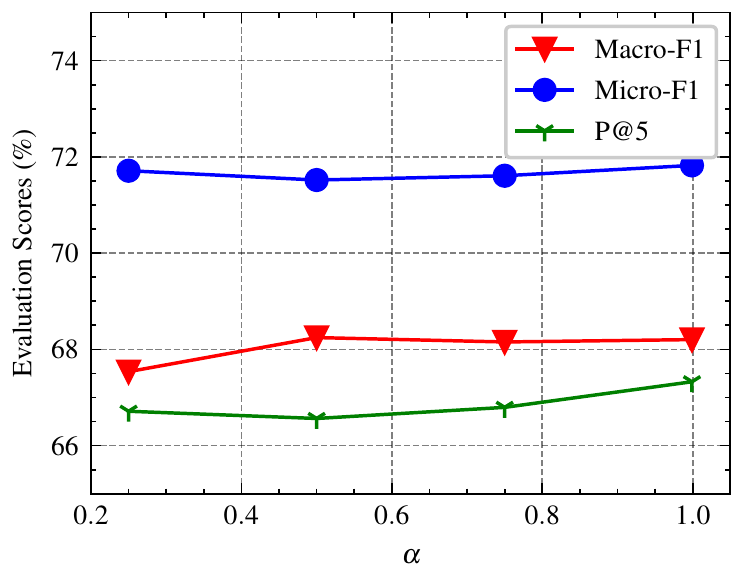}
    \caption{Evaluation results with $\gamma = 2$.}
    \label{fig:alpha}
\end{subfigure}
\quad
\begin{subfigure}[]{0.45\textwidth}
    \includegraphics[width=\textwidth]{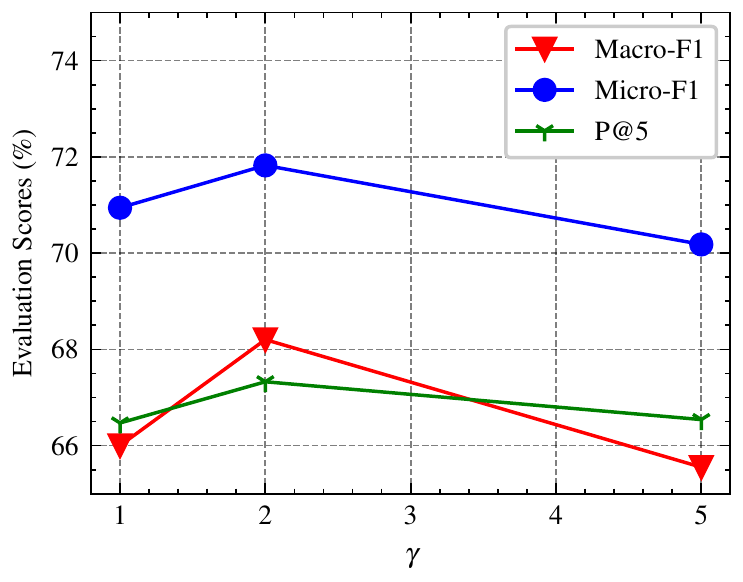}
    \caption{Evaluation results with $\lambda = 0.999$.}
    \label{fig:gamma}
\end{subfigure}
\caption{Predictive performance on MIMIC-III dataset with different $\gamma$ and $\lambda$.}
\label{fig:alpha_gamma}
\end{figure}

\textcolor{black}{
Figure~\ref{fig:alpha} shows that the predictive performance of the MARN on MIMIC-III dataset by applying different $\alpha$.
The variations of the MARN's performance is small, which means the adjustment of the hyper-parameter~$\alpha$ does not largely influence the evaluation results.
The oscillations of three scores are slight. 
We can find the optimal $\alpha$ is $0.999$.
Figure~\ref{fig:gamma} shows that $\gamma$ largely effects the results of our proposed model, and the optimal $\gamma$ is $2$.
Based on these two figures, we found that the $\gamma$ contributes more to balancing the learning of the high- and low-frequency codes and the performance increment of the parameter $alpha$ is slight.
}

\begin{figure}[htbp]
\centering
\begin{subfigure}[]{0.45\textwidth}
    \includegraphics[width=\textwidth]{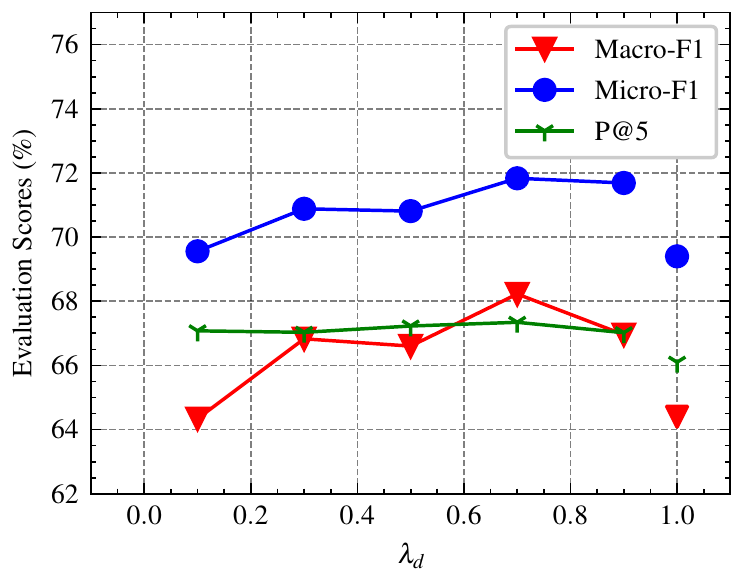}
    \caption{Evaluation results with different loss weight of ICD codes~($\lambda_d$).}
    \label{fig:lambda_d}
\end{subfigure}
\quad
\begin{subfigure}[]{0.45\textwidth}
    \includegraphics[width=\textwidth]{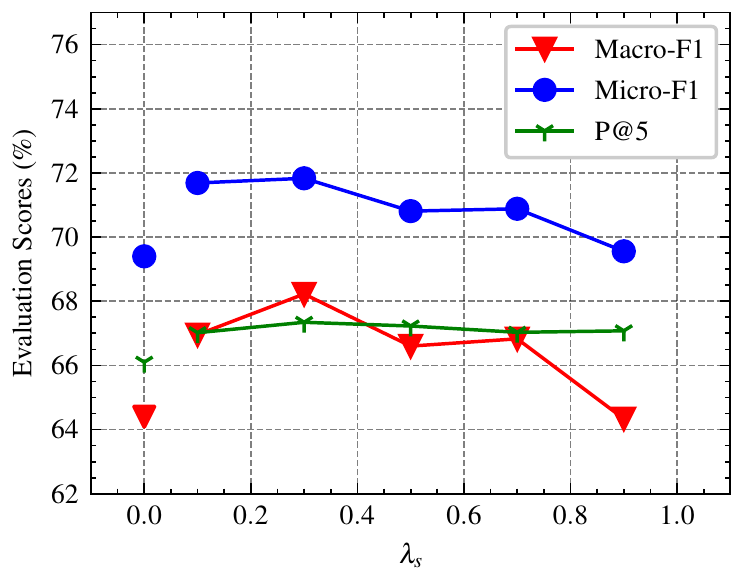}
    \caption{Evaluation results with different loss weight of CCS codes~($\lambda_s$).}
    \label{fig:lambda_s}
\end{subfigure}
\caption{Predictive performance on MIMIC-III dataset with different $\lambda_d$ and $\lambda_s$~($\lambda_d + \lambda_s = 1$).}
\label{fig:lambda_ds}
\end{figure}

\textcolor{black}{
We evaluate our proposed MARN on the MIMIC-III dataset by setting different loss weights of medical coding branches, i.e., $\lambda_d$ for the ICD coding and $\lambda_s$ for the CCS coding, with results shown in Figure~\ref{fig:lambda_d} and ~\ref{fig:lambda_s}, respectively.
Two figures are symmetrical because the summation of these two coefficients is $1$.
We take a common loss weighting strategy for our MTL scheme, which is to assign same weights for each coding tasks~\cite{lin2021closer}, and we fine-tune these loss weights to get the best performance.
Intuitively, the CCS coding branch as an auxiliary task should be assigned with smaller loss weight to enable the association knowledge transferred from the CCS coding branch to the ICD branch.
The variations of three scores are small in Figure~\ref{fig:lambda_d} when $\lambda_d$ is from $0.5$ to $0.9$.
However, if we further increase the $\lambda_d$ to $1$~(i.e., removing the MTL scheme from the MARN), the Macro-F1, Micro-F1, and P@5 dramatically drop to $64.4$, $69.4$, and $66.1$.
In summary, the evaluation results are not sensitive to the value of the $\lambda_d$ and $\lambda_s$ and the model's performance largely decreases if the MARN does not include the MTL scheme.
}

\section{Future work}
\label{sec:future}
In recent years, the transformer-based language model has become a new paradigm for NLP tasks. 
With the support of self-attention mechanism, the transformer-based models can capture token-dependent patterns for boosting contextualized text learning.
The transformer and its variants can suffer from the quadratic memory and time complexity problem caused by the self-attention mechanism.
Although the efficient transformer-based models, such as Reformer~\cite{kitaev2020reformer}, Linformer~\cite{wang2020linformer} Longformer~\cite{beltagy2020longformer},  have been proposed, they also need a substantial amount of computational resource for neural network training. 
By contrast, the GRU-based model is positioned nicely on the Pareto frontier of the computation-performance curve. 
In the next stage of our research, we plan to study how to effectively utilize contextual embeddings to obtain semantically enriched document features for medical code prediction. 

\section{Conclusion}
\label{sec:conclusion}
This paper proposes a novel model, \textbf{M}ultitask b\textbf{A}alanced and \textbf{R}ecalibrated \textbf{N}etwork~(MARN), to tackle three challenges of automated medical coding: the imbalanced class problem, capturing code association, and dealing with lengthy and noisy documents.
We leverage the focal loss to alleviate the imbalanced class issue by redistributing the loss weights between low and high-frequency medical codes.
We design the \textbf{R}ecalibrated \textbf{A}ttention \textbf{M}odule~(RAM) to inject high-level semantic features into the original feature for noise suppressing. 
The cascaded convolutional structure of the RAM can improve the representation learning from long and noisy documents. 
The multitask learning scheme that enables the code relationship knowledge transfer between two different coding systems (i.e., ICD and CCS) is developed to capture the code association and improve the coding performance.
The experimental results show that our proposed model outperforms competitive baseline models in the real-world MIMIC-III database.

\section{Acknowledgments}

This work was supported by the Academy of Finland (grant 336033) and EU H2020 (grant 101016775).
We acknowledge the computational resources provided by the Aalto Science-IT project.
The authors wish to acknowledge CSC - IT Center for Science, Finland, for computational resources.

\bibliographystyle{ACM-Reference-Format}
\bibliography{MARN_medical_coding}

\end{document}